\documentclass[10pt,twocolumn,letterpaper]{article}

\usepackage[pagenumbers]{wacv} %

\usepackage{graphicx}
\usepackage{amsmath}
\usepackage{amssymb}
\usepackage{pifont}
\usepackage{booktabs}
\usepackage[dvipsnames,]{xcolor}
\usepackage{url}
\usepackage{graphicx, subcaption}
\usepackage[export]{adjustbox}
\usepackage{pgffor}
\usepackage[accsupp]{axessibility}  %
\usepackage{titletoc}
\usepackage{enumitem}
\usepackage{siunitx}
\usepackage{multirow}
\usepackage{orcidlink}          %
\usepackage[hang,symbol]{footmisc}

\newcommand{\std}[1]{{\fontsize{6}{7.2}\selectfont ±#1}} %
\newcommand{\stdt}[1]{{\fontsize{5}{7.2}\selectfont ±#1}}

\newcommand{\normfixed}[1]{\lVert#1\rVert}

\usepackage[capitalize]{cleveref}
\crefname{section}{Sec.}{Secs.}
\Crefname{section}{Section}{Sections}
\Crefname{table}{Table}{Tables}
\crefname{table}{Tab.}{Tabs.}
\crefname{theorem}{Theorem}{Theorems}
\crefname{theorem}{Thm.}{Thms.}

\def\wacvPaperID{548} %
\def\confName{WACV}
\def\confYear{2025}

\newcommand{\lossmse}{\ensuremath{\mathcal{L}_{\text{MSE}}}}
\newcommand{\lossLone}{\ensuremath{\mathcal{L}_{\text{L1}}}}
\newcommand{\losslogLone}{\ensuremath{\mathcal{L}_{\text{logL1}}}}

\newcommand{\inst}[1]{{$^{#1}$}}

\begin{document}

\title{RAW-Diffusion: RGB-Guided Diffusion Models\\for High-Fidelity RAW Image Generation}

\author{
    \begin{NoHyper}
    Christoph Reinders\inst{1}\footnotemark[2]\quad  %
    Radu Berdan\inst{2}\footnotemark[1] \quad
    Beril Besbinar\inst{2}\footnotemark[1] \quad
    Junji Otsuka\inst{3} \quad
    Daisuke Iso\inst{2}
    \end{NoHyper}
    \\
    \inst{1}Leibniz University Hannover \quad
    \inst{2}Sony AI \quad
    \inst{3}Sony Group Corporation\\
}

\maketitle

\footnotetext[2]{Work done during an internship at Sony AI. Corresponding author: \tt{reinders@tnt.uni-hannover.de}\label{fnlabel}}
\footnotetext[1]{These authors contributed equally to this work}

\begin{abstract}
Current deep learning approaches in computer vision primarily focus on RGB data sacrificing information. In contrast, RAW images offer richer representation, which is crucial for precise recognition, particularly in challenging conditions like low-light environments. 
The resultant demand for comprehensive RAW image datasets contrasts with the labor-intensive process of creating specific datasets for individual sensors.
To address this, we propose a novel diffusion-based method for generating RAW images guided by RGB images.
Our approach integrates an RGB-guidance module for feature extraction from RGB inputs, then incorporates these features into the reverse diffusion process with RGB-guided residual blocks across various resolutions. 
This approach yields high-fidelity RAW images, enabling the creation of camera-specific RAW datasets.
Our RGB2RAW experiments on four DSLR datasets demonstrate state-of-the-art performance.
Moreover, RAW-Diffusion demonstrates exceptional data efficiency, achieving remarkable performance with as few as $25$ training samples or even fewer.
We extend our method to create BDD100K-RAW and Cityscapes-RAW datasets, revealing its effectiveness for object detection in RAW imagery, significantly reducing the amount of required RAW images. 
The code is available at 
\textcolor{magenta}{\url{https://github.com/SonyResearch/RAW-Diffusion}}.
\end{abstract}

\section{Introduction}
\label{sec:intro}

In the traditional imaging pipeline, a camera sensor captures a RAW image, which is then converted via an Image Signal Processor (ISP) and compressed to a RGB image tailored for human consumption. 
However, while RGB images are convenient for human perception and readily available, their processing from RAW format introduces artifacts and information loss. 
Nonetheless, RGB images are ubiquitous in contemporary computer vision applications and serve as the primary data source for training neural networks. 
When running large computer vision models in the cloud, performing inference on RGB images is advantageous for reasons other than model performance: they are more cost-effective to transmit and store than their original RAW images.
On the other hand, when running CV models on the edge, under strict resource and hardware constraints, extracting the maximum information from the scene and avoiding unnecessary pre-processing in order to maximize small model performance becomes critical.

\begin{figure}[t]
    \centering
    \includegraphics[width=\columnwidth]{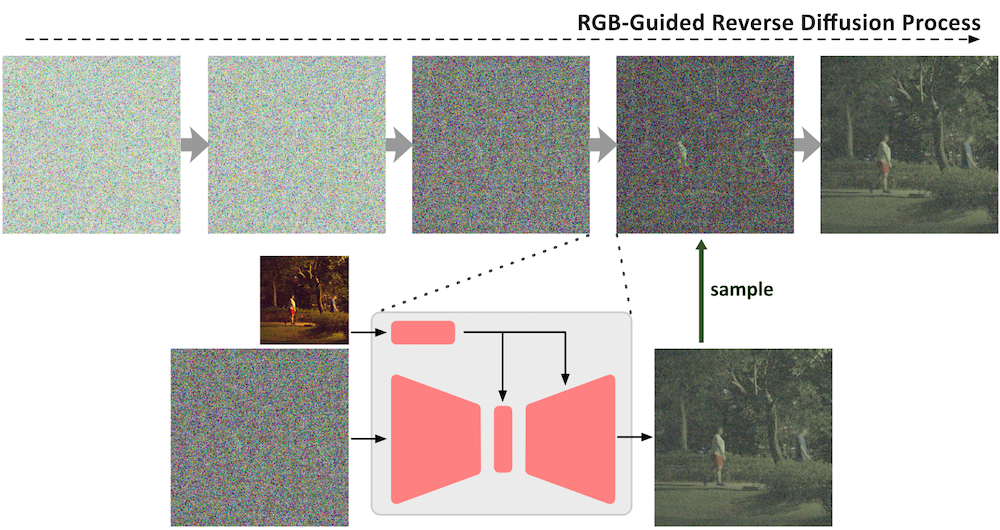}
    \caption{
    RAW-Diffusion enables the generation of high-fidelity RAW images by iterative denoising of a noisy RAW input through a RGB-guidance module and RGB-guided residual blocks.
    RAW-Diffusion is the first successful diffusion-based method for RAW generation that outperforms state-of-the-art methods.
    }
    \vspace{-0.2cm}
    \label{fig_teaser}

\end{figure}

The limitations of RGB images are particularly evident in scenarios requiring precise scene irradiance information, such as low-light environments or fine-grained image manipulation tasks.
In contrast, RAW images captured directly by camera sensors offer a more faithful representation of scene data, retaining higher quantization levels and preserving the unprocessed sensor readings. 
The enhanced dynamic range of RAW images has demonstrated superior performance for a variety of image editing and computer vision tasks, including photometric stereo~\cite{shi2016benchmark}, denoising~\cite{gharbi2016deep,abdelhamed2018high,brooks2019unprocessing}, reflection removal~\cite{lei2023robust}, low-light image enhancement~\cite{dagli2023diffuseraw}, image super-resolution~\cite{zhang2019zoom}, and object detection~\cite{otsukaSelfSupervisedReversedImage2023}. 
Employing RAW image inputs can thus improve the accuracy and robustness of neural networks across various computer vision tasks~\cite{buckler2017reconfiguring}, especially when running on the edge. 
However, sensor-specific data collection and annotation for RAW datasets is extremely costly. 
Moreover, accessing RAW images can be challenging due to their memory-intensive nature, posing difficulties in storage, transfer, and sharing.

\parskip=0pt
To address this issue, several approaches have been proposed to map RGB images to corresponding RAW sensor outputs~\cite{mitsunaga1999radiometric,afifi2021cie,grossberg2003determining,grossberg2003determining,lin2005determining,chakrabarti2009empirical,kim2012new,xingInvertibleImageSignal2021,zamir2020cycleisp,condeReversedImageSignal2022}, some relying on metadata or prior camera information for reverse ISP reconstruction~\cite{mitsunaga1999radiometric, grossberg2003determining,lin2005determining,debevec2023recovering,brooks2019unprocessing}, which is not always accessible.
Recent approaches aim to learn invertible functions for mapping between RGB and RAW images, achieving superior RAW synthesis performance.
Among those, many methods~\cite{liang2021cameranet,liu2022deep,yoshimura2023dynamicisp} mimic traditional ISP pipelines, requiring sensor-specific configurations but being more data efficient. 
Others~\cite{zamir2020cycleisp,xingInvertibleImageSignal2021,brooks2019unprocessing} learn nonlinear mappings using RGB-RAW image pairs, often requiring many training samples.
Our goal is to develop a flexible, data-efficient, and high-performing method for RAW image synthesis.

In this work, we present a novel diffusion model for RAW image reconstruction, employing features derived from an RGB image (see \cref{fig_teaser}).
Unlike conventional methods that simply concatenate or add guidance signals \cite{liuInvertingImageSignal2022}, we introduce RGB-guided residual blocks for integrating RGB features and guiding the denoising of the RAW images.
This approach not only achieves state-of-the-art performance in RAW image reconstruction but also facilitates a data-efficient method for generating large-scale sensor-specific datasets using existing RGB images. 
In summary, our \textbf{contributions} are as follows:
\begin{itemize}
    \setlength\itemsep{0pt}
    \item We propose a novel diffusion model for accurate RAW image reconstruction with RGB image guidance.
    \item The proposed method demonstrates superior performance in RAW image synthesis with high-bit sensors, especially for images featuring challenging irradiance distributions, and surpasses state-of-the-art methods.
    \item Our method is extremely data-efficient, achieving the same performance with $25$ or fewer training samples as when training on the full training set.
    \item We enable the creation of camera-specific RAW datasets from existing RGB datasets, eliminating the need for costly sensor-specific data acquisition.
    \item The proposed method bridges the gap between models trained on RGB images for downstream tasks and RAW images, as shown for object detection.
\end{itemize}

\section{Related Work}

\paragraph{RAW Image Reconstruction.} 
Unmapping the processed RGB images back to the RAW domain has been a longstanding research topic. 
Conventional radiometric calibration algorithms use multiple images taken with controlled exposures to compute the response function of a camera’s output intensity values with respect to the amount of light falling on the sensor~\cite{mitsunaga1999radiometric, grossberg2003determining,lin2005determining,debevec2023recovering}, which are further advanced to handle saturated RGB values~\cite{chakrabarti2009empirical,kim2012new,chakrabarti2014modeling}.
However, these methods require calibration for a given camera, which involves multiple parameterized models for different camera settings.
Modern deep learning-based algorithms~\cite{afifi2021cie,brooks2019unprocessing,conde2022model,condeReversedImageSignal2022,gharbi2016deep,schwartz2018deepisp,liu2022deep,zamir2020cycleisp,xingInvertibleImageSignal2021,liang2021cameranet,liu2022deep} have shown how data-driven approaches can outperform conventional methods without the heavy need for calibration.
These methods either replace each or groups of ISP elements with neural networks~\cite{liang2021cameranet,liu2022deep,yoshimura2023dynamicisp} or model RGB-to-RAW or RAW-to-RGB mapping with a single network~\cite{zamir2020cycleisp,xingInvertibleImageSignal2021,brooks2019unprocessing}.
While being more data efficient, the first approach requires sensor-specific configuration.
In contrast, the single network approach requires large datasets of RGB-RAW image pairs.

\vspace{-0.2cm}
\paragraph{Diffusion Models.}
Diffusion models~\cite{hoDenoisingDiffusionProbabilistic2020}, a probabilistic generative approach, have recently gained prominence for their ability to produce high-quality images with diverse features. 
They operate through a forward process, adding noise to clean samples, and a reverse process, recovering samples from noise~\cite{luo2022understanding,croitoru2023diffusion}. 
Offering stable training and strong priors compared to other generative models like GANs and VAEs~\cite{hoDenoisingDiffusionProbabilistic2020}, these models excel in various tasks like image generation~\cite{ho2022cascaded,dhariwal2021diffusion,nicholGLIDEPhotorealisticImage2022,rombach2022high,ramesh2022hierarchical}, inpainting~\cite{lugmayr2022repaint,xie2023smartbrush}, molecular
generation~\cite{xuGeoDiffGeometricDiffusion2022}, audio synthesis~\cite{kongDiffWaveVersatileDiffusion2021}, and medical image analysis~\cite{kazerouniDiffusionModelsMedical2023}.
Denoising diffusion probabilistic models (DDPMs)~\cite{hoDenoisingDiffusionProbabilistic2020}, primarily focus on removing noise from images, while Denoising Diffusion Implicit Models (DDIMs)~\cite{songDenoisingDiffusionImplicit2021} generalize DDPMs in a more computationally efficient inference scheme via a non-Markovian forward process.
As a generative model, they show the potential for enhancing datasets to improve downstream tasks like image classification~\cite{trabuccoEffectiveDataAugmentation2024} or object detection~\cite{fang2024data}.

\vspace{-0.2cm}
\paragraph{Object detection.}
Deep learning methods have demonstrated remarkable performance across various domains of computer vision \cite{krizhevskyImageNetClassificationDeep2012,heDeepResidualLearning2016,zhangMixupEmpiricalRisk2018,dosovitskiyImageWorth16x162021,caoTrainingVisionTransformers2022,reindersChimeraMixImageClassification2022,heMaskedAutoencodersAre2022,reindersTwoWorldsOne2024,oquabDINOv2LearningRobust2023}, including significant improvements in object detection through data-driven approaches~\cite{zou2023object,carionEndendObjectDetection2020,redmon2016you,renFasterRCNNRealtime2015,reindersObjectRecognitionVery2018,reindersLearningConvolutionalNeural2019,liuGroundingDINOMarrying2024}.
Object detection methods are generally classified into two categories: single-stage~\cite{sermanet2013overfeat,liuSSDSingleShot2016,lin2017focal,redmon2016you,tian2019fcos,tan2020efficientdet} and two-stage methods approaches~\cite{girshick2014rich,lin2017feature,renFasterRCNNRealtime2015}.
Notably, models like the YOLO~\cite{redmon2016you} family and Faster R-CNN~\cite{renFasterRCNNRealtime2015} have been widely adopted as representative approaches.
Recently, there has been a growing interest in leveraging RAW images for object detection~\cite{buckler2017reconfiguring,morawskiGenISPNeuralISP2022,yoshimura2023dynamicisp,yoshimura2023rawgment,ljungbergh2023raw,hong2021crafting}, as they offer more precise information, particularly beneficial in low-light environments~\cite{hong2021crafting}.

\section{Method}
The proposed diffusion-based method is able to iteratively generate high-fidelity RAW images, simulate the distribution of the camera sensor, and create specific scenes by injecting RGB-guidance at multiple resolutions with RGB-guidance residual blocks into the diffusion process. 
In the following, we will briefly present the foundations of DDPMs. 
Afterward, the RGB-guidance module is introduced, which is followed by the architecture of the diffusion model including the conditional image generation. 
Finally, the training objective of the diffusion model is presented.

\subsection{Preliminaries} 
Diffusion models \cite{sohl-dicksteinDeepUnsupervisedLearning2015,hoDenoisingDiffusionProbabilistic2020} are latent variable generative models inspired by non-equilibrium thermodynamics.
In the forward diffusion process, a Markov chain of $T$ steps is formulated by gradually adding random noise to the data.
Given a data sample from the real data distribution $x_0 \sim q(x)$, a sequence of latent variables $x_1, \dots, x_T$ is generated by progressively adding Gaussian noise:%
\begin{equation}
    q(x_t \mid x_{t-1}) = \mathcal{N}(x_t; \sqrt{1-\beta_t}x_{t-1}, \beta_tI), %
\end{equation}
where $0 < \beta_t < 1$ is the variance schedule of the Gaussian noise, and $I$ the identity matrix.
By defining $\alpha_t = 1 - \beta_t$ and $\bar{\alpha}_t = \prod_{i=1}^t \alpha_t$, $x_t$ can be sampled at an arbitrary time step in closed form using the reparametrization trick $x_t = \sqrt{\bar{\alpha}_t}x_0 + \sqrt{1-\bar{\alpha}}_t\epsilon$ with $\epsilon \in \mathcal{N}(0, I)$, i.e., 
$q(x_t\mid x_0) = \mathcal{N}(x_t; \sqrt{\bar{\alpha}_t}x_0, \sqrt{1-\bar{\alpha}}_tI)$.

The reverse diffusion process is learned by training a parameterized model $p_{\theta}$ to recover the data and gradually remove noise.
Starting from random noise $x_T \sim \mathcal{N}(0, I)$, the distribution of $x_{t-1}$ given $x_{t}$ and $t$ is modeled as
\begin{equation}
    p_{\theta}(x_{t-1} \mid x_t) = \mathcal{N}(x_{t-1};\mu_{\theta}(x_t, t), \sigma^2_tI)
\end{equation}
to generate new data samples, where $\mu_{\theta}$ is the Gaussian mean estimated by the denoising network and $\sigma_t$ is a fixed variance.

\begin{figure}[t]
    \centering
    \includegraphics[width=\columnwidth]{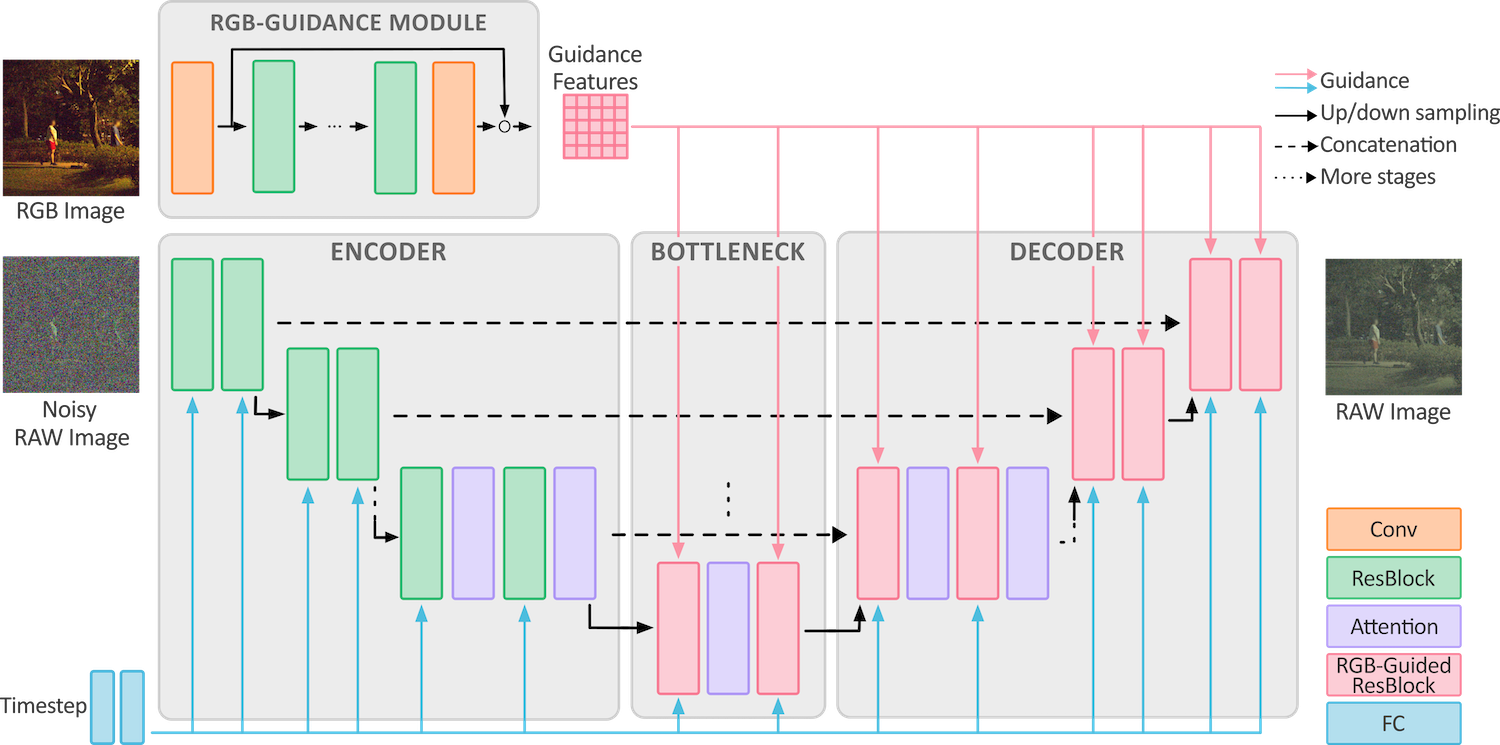}
    \caption{The RAW-Diffusion architecture consists of an RGB-guidance module for creating guidance features and an encoder for processing noisy RAW inputs. The guidance features are then integrated into both the bottleneck and decoder with RGB-guidance residual blocks, modulating the diffusion features to reconstruct the RAW image.
    }
    \vspace{-0.2cm}
    \label{fig_overview}
\end{figure}

\subsection{RAW-Diffusion}
RAW-Diffusion is based on DDPMs and learns the reconstruction of a RAW image by starting from random noise and iterative denoising of a noisy RAW image guided by an RGB image. 
We introduce an RGB-guidance module that extracts features from the RGB images. 
The diffusion model, utilizing a U-Net architecture, injects the features into the diffusion process to guide the generation, and directly predicts the RAW image, without $\epsilon$-prediction. 
An overview of RAW-Diffusion is shown in \cref{fig_overview} and the individual components are introduced in the following sections.

\vspace{-0.2cm}
\subsubsection{RGB-Guidance Module}
For high-fidelity RAW image synthesis, the guidance features from the ISP-processed RGB image $I_{\text{RGB}} \in \mathcal{R}^{3 \times H \times W}$ should contain both low- and high-level information.
Hence, we propose using the enhanced deep super-resolution network (EDSR)~\cite{limEnhancedDeepResidual2017} without an upsampling head for feature extraction. 
The RGB image $I_{\text{RGB}}$ is first processed by a convolutional layer to lift the RGB channels to $C_{\text{GM,Features}}$ channels, which are later fed to a sequence of $N_{\text{GM,ResBlocks}}$ residual blocks.
Each residual block consists of two convolutional layers followed by ReLU activation and a skip connection. 
The output of the residual blocks is then added to the initial convolved features, resulting in a feature map $F_{\text{RGB}} \in \mathcal{R}^{C \times H \times W}$, where $C=C_{\text{GM,Features}}$.
All convolutional layers employ  $3 \times 3$ kernels with reflective padding of size 1, and the guidance module does not contain any normalization layers based on empirical results.

\vspace{-0.2cm}
\subsubsection{Encoder}

The encoder network follows a standard architecture~\cite{dhariwalDiffusionModelsBeat2021} and derives a multi-scale feature representation of the noisy RAW image.
The input at each level is processed by $N_{\text{ResBlocks}}$ residual blocks, followed by a downsampling layer that halves the spatial resolution, while expanding the feature dimension.
Each residual block is characterized by two convolutional layers, each followed by SiLU activation ~\cite{hendrycksGaussianErrorLinear2016,elfwingSigmoidWeightedLinearUnits2018,ramachandranSearchingActivationFunctions2018} and group normalization \cite{wuGroupNormalization2018}.
The current timestep $t$ of the diffusion process is mapped to a learnable embedding that is later projected onto a set of affine parameters for channel-specific transformation of the encoder features after the second group normalization.
We also employ self-attention blocks after each residual block at specific resolutions to facilitate learning long-term dependencies between different spatial regions.

\vspace{-0.2cm}
\subsubsection{RGB-Guided Residual Block}
Our objective is the guidance of the diffusion process for reconstructing RAW images using the relevant RGB information. 
For an overview of the fundamentals of guiding diffusion models, please refer to \cite{ho2021classifierfree,zhangAddingConditionalControl2023,mouT2IAdapterLearningAdapters2024}.
A naive solution involves concatenating the additional information directly with the noisy input. 
However, we introduce an RGB-guided residual block to inject RGB information into the network, enhancing the RAW image reconstruction. 
For that, we integrate the RGB-guidance features by modulating the feature transformation within the residual block.
The RGB-guided residual block is inspired by the semantic diffusion model \cite{wangSemanticImageSynthesis2022} which uses spatially-adaptive normalization (SPADE) \cite{parkSemanticImageSynthesis2019} to condition the diffusion process.
We introduce two layer-specific modules $\gamma^i$ and $\beta^i$ which downscale the RGB guidance features by bilinear interpolation and predict modulation parameters $\gamma^i(x) \in \mathcal{R}^{C \times H' \times W'}$ and $\beta^i(x) \in \mathcal{R}^{C \times H' \times W'}$ for each feature and spatial position, where $H' \times W'$ is the size of the feature map and $C$ number of channels. 
The modules $\gamma^i$ and $\beta^i$ consist of a shared convolutional layer with ReLU activation followed by another individual convolutional layer.
Afterward, the RGB information is integrated by scaling and shifting the normalized diffusion features $F^i$ as follows:
\begin{equation}
    F^{i+1} =  \operatorname{Norm}(F^{i}) \cdot (1 + \gamma^i(\operatorname{Down}(F_{\text{RGB}}))) + \beta^i(\operatorname{Down}(F_{\text{RGB}})).
\end{equation}
The RGB-guided spatially adaptive normalization envelops both instances of group normalization within each residual block, learning to incorporate the relevant RGB information across multiple resolutions to guide the RAW image reconstruction process effectively.

\vspace{-0.2cm}
\subsubsection{Bottleneck}
Starting with the bottleneck, we inject the RGB-guidance features into the diffusion process by modulating the high-level diffusion features with the proposed RGB-guided residual block. 
The bottleneck is composed of an RGB-guided ResBlock, an attention block, and another RGB-guided ResBlock.

\vspace{-0.2cm}
\subsubsection{Decoder}

The decoder, symmetric to the encoder, takes the concatenation of the upscaled features from the preceding level and the shortcut connection from the encoder as input at each layer.
Each decoder level is composed of a series of RGB-guided residual blocks and attention blocks.
The final level decoder features are then projected via a convolutional layer and a hyperbolic tangent function to compose the RAW image.
Our method directly predicts the denoised RAW image rather than estimating noise.

\subsection{Training Objective}
The diffusion model is trained with pairs of RAW and RGB images, denoted as $I_{\text{RAW}}$ and $I_{\text{RGB}}$
In each training step, noisy RAW images and RGB images are provided as input, and the model is optimized to directly predict the RAW image $\hat{I}_{\text{RAW}}$.
We define three loss functions to facilitate the accurate reconstruction of the original RAW image.
First, the squared difference and mean absolute difference are optimized via $\lossmse$ and $\lossLone$, respectively.
In addition, to address the bias in the distribution of the RAW pixel values towards low values, a logarithmic L1 loss \cite{eilertsenHDRImageReconstruction2017}, defined as $ \losslogLone(\hat{I}, I) = \frac{1}{CHW} \normfixed{\log(\hat{I} + \epsilon) - \log(I+ \epsilon)}_1$, where $\epsilon$ serves as a minimal constant to ensure numerical stability, is adopted.
For the $\losslogLone$, the value range is adjusted to $[0, 1]$.
Hence, the overall training objective is defined as the composition of all three losses: $\mathcal{L} = \mathcal{L}_{\text{MSE}} + \mathcal{L}_{\text{L1}} + \mathcal{L}_{\text{logL1}}$.

\section{Experiments}
\label{sec_experiments}
We perform extensive experiments to evaluate the performance of RAW-Diffusion and state-of-the-art methods. 

\subsection{Datasets}
The experiments are performed on diverse image sets captured by four different DSLR cameras, two from each the MIT-Adobe FiveK~\cite{bychkovskyLearningPhotographicGlobal2011} and NOD datasets~\cite{morawskiGenISPNeuralISP2022}.
Following \cite{xingInvertibleImageSignal2021}, we use 777 and 590 images captured by the Canon EOS 5D camera and Nikon D700 cameras, respectively, and use a train-test split of $85/15$.
The Night Object Detection (NOD) dataset \cite{morawskiGenISPNeuralISP2022} consists of RAW images recorded by Sony RX100 VII and Nikon D750, with object annotations of three classes.
The Sony camera has 3.2k images with 18.7k annotated instances, and Nikon has 4.0k with 28.0k annotated instances. We use the official training and test split.
All RAW images are processed using the RawPy library to generate the corresponding RGB image.
Additionally, we integrate two large-scale RGB datasets: Cityscapes \cite{cordtsCityscapesDatasetSemantic2016} and BDD100K \cite{Yu2018BDD100KAD} (denoted as CS and BDD). Cityscapes consists of 5,000 images with pixel-level instance segmentation annotations recorded in street scenes from 50 different cities. BDD100K is a diverse dataset with 100k videos and various tasks.
The dataset provides object detection annotations for 100k keyframes.

\begin{table*}[t]
  \caption{Quantitative RGB2RAW results on the test set of four DSLR cameras evaluating the RAW reconstruction performance using PSNR ($\uparrow$) and SSIM ($\uparrow$) of various state-of-the-art methods and RAW-Diffusion. We report the mean performance and standard deviation. The best result is highlighted in bold, and the second best underlined.\vspace{-0.1cm}}
  \small
  \label{tab:exp_results_rgb2raw}
  \centering
  \begin{tabular}{l|
  cc|
  cc|
  cc|
  cc
  }
    \toprule
           & \multicolumn{2}{c}{\textbf{FiveK Nikon}} & \multicolumn{2}{c}{\textbf{FiveK Canon}} &\multicolumn{2}{c}{\textbf{NOD Nikon}} & \multicolumn{2}{c}{\textbf{NOD Sony}} \\
    Method & PSNR & SSIM & PSNR & SSIM & PSNR & SSIM & PSNR & SSIM\\
    \midrule

U-Net \cite{condeReversedImageSignal2022}
    & 26.00\std{1.02}  & 0.507\std{0.06}  
    & 27.77\std{2.23}  & 0.691\std{0.09}    
    & 37.90\std{2.19}  & 0.791\std{0.10}   
    & 37.71\std{0.30}  & 0.813\std{0.02} \\

UPI \cite{brooksUnprocessingImagesLearned2019}
    & 27.66\std{0.03}  & 0.830\std{0.00} 
    & 30.68\std{0.08}  & 0.893\std{0.00} 
    & 37.12\std{0.05}  & 0.816\std{0.00} 
    & 33.49\std{0.02}  & 0.735\std{0.00} \\

InvGrayscale \hspace{-1pt}\cite{xiaInvertibleGrayscale2018}
    & 28.32\std{0.55}  & 0.838\std{0.01} \
    & 29.67\std{0.43}  & 0.884\std{0.00} 
    & 30.41\std{0.38}  & 0.803\std{0.01} 
    & 29.52\std{0.42}  & 0.756\std{0.00} \\

InvISP \cite{xingInvertibleImageSignal2021} 
    & 26.41\std{0.19}   & 0.732\std{0.01} 
    & 30.26\std{0.42}   & 0.885\std{0.00} 
    & 32.50\std{1.91}   & 0.844\std{0.05} 
    & 23.35\std{3.50}   & 0.533\std{0.11} \\

InvISP$^{+}$ \cite{xingInvertibleImageSignal2021} 
    & 28.95\std{0.15}   & 0.841\std{0.00} 
    & \underline{32.23}\std{0.09}   & \underline{0.912}\std{0.00} 
    & \underline{42.57}\std{0.09}   & \underline{0.973}\std{0.00} 
    & 38.27\std{0.01}   & 0.916\std{0.00} \\

ISPLess \cite{dattaEnablingISPlessLowpower2023}
    & 26.82\std{0.15}   & 0.762\std{0.02} 
    & 29.73\std{1.44}   & 0.880\std{0.01} 
    & 33.04\std{0.71}   & 0.870\std{0.02} 
    & 29.66\std{0.39}   & 0.669\std{0.03} \\

ISPLess$^{+}$ \cite{dattaEnablingISPlessLowpower2023}
    & \underline{29.83}\std{0.41}   & \underline{0.845}\std{0.00} 
    & 32.15\std{0.25}   & 0.904\std{0.00} 
    & 41.50\std{0.02}   & 0.967\std{0.00} 
    & 38.31\std{0.07}   & \underline{0.917}\std{0.00} \\

CycleR2R \cite{liEfficientVisualComputing2024}
    & 28.40\std{0.85}   & 0.842\std{0.01} 
    & 24.85\std{3.45}   & 0.791\std{0.05} 
    & 24.93\std{1.86}   & 0.562\std{0.03} 
    & 23.71\std{1.4}    & 0.491\std{0.02} \\

RISPNet \cite{dongRISPNetNetworkReversed2023}
    & 29.00\std{NA}    & 0.669\std{NA}     
    & 29.56\std{NA}    & 0.789\std{NA}   
    & 40.07\std{NA}    & 0.898\std{NA}   
    & \underline{38.67}\std{NA}    & 0.871\std{NA} \\

Diffusion \cite{liuInvertingImageSignal2022}
    & 27.64\std{0.07} & 0.798\std{0.00} & 29.17\std{0.15} & 0.840\std{0.00} & 35.59\std{1.29} & 0.867\std{0.02} & 33.07\std{1.16} & 0.812\std{0.02} \\

SRISP \cite{otsukaSelfSupervisedReversedImage2023}
    & 28.19\std{0.15}  & 0.797\std{0.00} 
    & 30.34\std{0.27}  & 0.844\std{0.00} 
    & 38.23\std{0.85}  & 0.880\std{0.01} 
    & 35.22\std{0.76}  & 0.815\std{0.01} \\

RAW-Diffusion (ours)
    & \textbf{30.05}\std{0.32}   & \textbf{0.857}\std{0.00} 
    & \textbf{34.01}\std{0.27}   & \textbf{0.926}\std{0.00} 
    & \textbf{44.93}\std{0.15}   & \textbf{0.983}\std{0.00} 
    & \textbf{39.17}\std{0.03}   & \textbf{0.932}\std{0.00} \\
\bottomrule  
\end{tabular}
\end{table*}

\subsection{Experimental Setup}
\paragraph{Training Details.}
We normalize the data to the range $[-1,1]$. The RAW images are scaled based on the black and white level.
During training, patches of size $256 \times 256$ are sampled by using random cropping, rotation, and flipping augmentation.
RAW-Diffusion is trained for 70k steps using AdamW \cite{loshchilovDecoupledWeightDecay2019} with an initial learning rate of $0.0001$, no weight decay, and a batch size of $4$. The learning rate is decreased linearly to zero.
Following \cite{nicholImprovedDenoisingDiffusion2021}, the number of diffusion steps, T, is set to 1000, and a linear variance schedule is used ranging from $\beta_1 = 0.0001$ to $\beta_T = 0.02$.
For object detection, Faster R-CNN \cite{renFasterRCNNRealtime2015} and YOLOv8 \cite{yolov8} models are trained. Details and other hyper-parameters are provided in the supplementary material.
All trainings are conducted on a single NVIDIA Tesla V100 with 16 GB memory, and repeated with three different seeds. 

\vspace{-0.2cm}
\paragraph{Evaluation.}
In line with prior works, we evaluate the PSNR and SSIM \cite{wangImageQualityAssessment2004} to assess the quality of the generated RAW images.
The images are normalized to the range of $[0, 1]$ for calculating the metrics.
Similar to \cite{xingInvertibleImageSignal2021}, we use a patch-based evaluation on a regular $3 \times 3$ grid. 
The precision of the object detection models is analyzed using the Average Precision (AP) \cite{linMicrosoftCOCOCommon2014}. 

\subsection{RGB2RAW Results}
We evaluate the RAW reconstruction performance of our proposed RAW-Diffusion and various state-of-the-art methods such as U-Net \cite{condeReversedImageSignal2022}, UPI \cite{brooksUnprocessingImagesLearned2019}, InvGrayscale \cite{xiaInvertibleGrayscale2018}, InvISP \cite{xingInvertibleImageSignal2021}, InvISP$^{+}$ \cite{xingInvertibleImageSignal2021}, ISPLess \cite{dattaEnablingISPlessLowpower2023}, ISPLess$^{+}$ \cite{dattaEnablingISPlessLowpower2023}, CycleR2R \cite{liEfficientVisualComputing2024}, RISPNet \cite{dongRISPNetNetworkReversed2023}, Diffusion \cite{liuInvertingImageSignal2022}, and SRISP \cite{otsukaSelfSupervisedReversedImage2023}. 
For UPI, the model parameters are optimized using the camera metadata, as described in the paper.
It should be noted that we are able to reproduce the InvISP results reported by the authors. However, the official implementation evaluates the inverse pass by rendering the sRGB images through the forward pass of the invertible neural network.
Here, we use the same RGB images for all methods.
We improve the performance of InvISP and ISPLess by removing the JPEG simulation and training the inverse process with the ground truth RGB image as input, denoted as InvISP$^{+}$ and ISPLess$^{+}$.
For RISPNet, we do not perform either model or test-time ensembling. In addition, we perform a single training due to computational requirements despite the downscaled model size.
For SRISP, the mean global feature of all training images is used as test-time reference which has proven more effective than sampling a reference.
Diffusion is trained on $64 \times 64$ patches as described by the authors. The performance with larger patch sizes collapses.

\begin{figure}[t]
    \centering
    \includegraphics[width=\columnwidth]{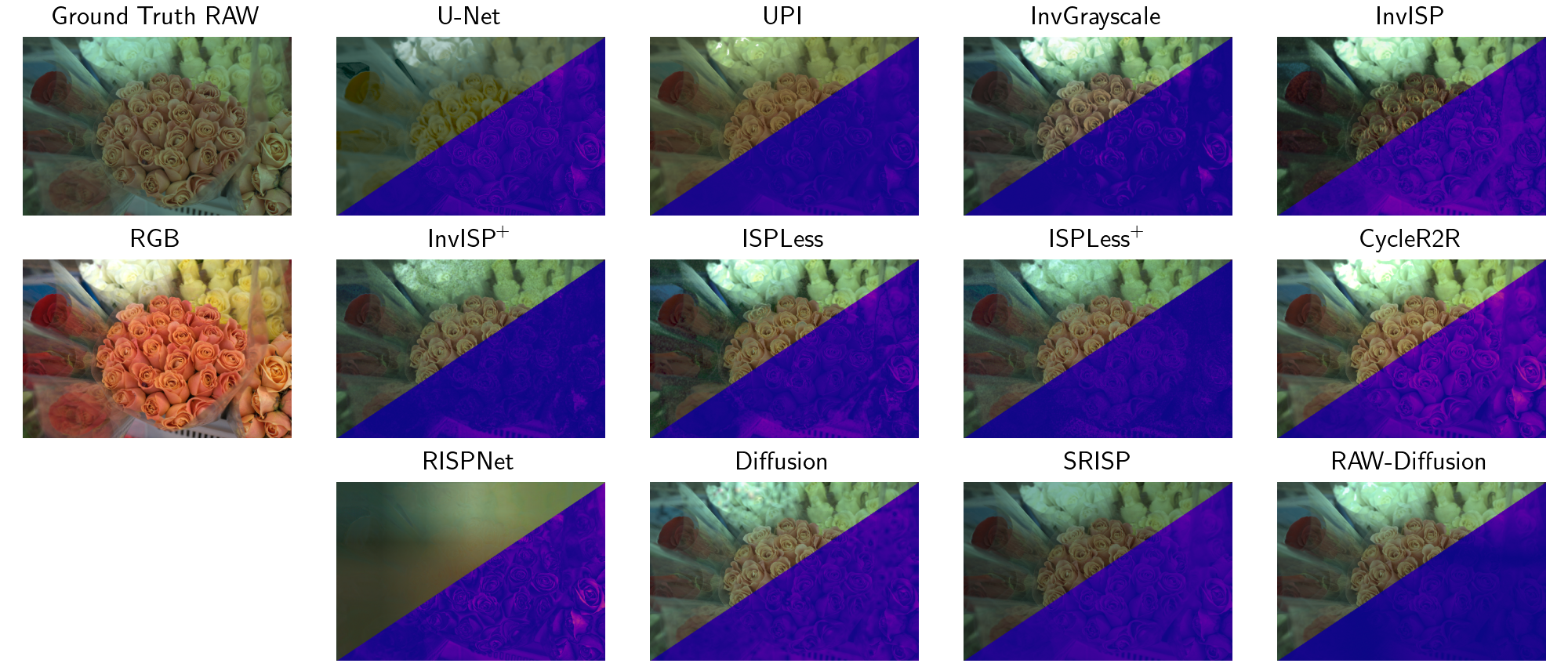}
    \includegraphics[width=\columnwidth]{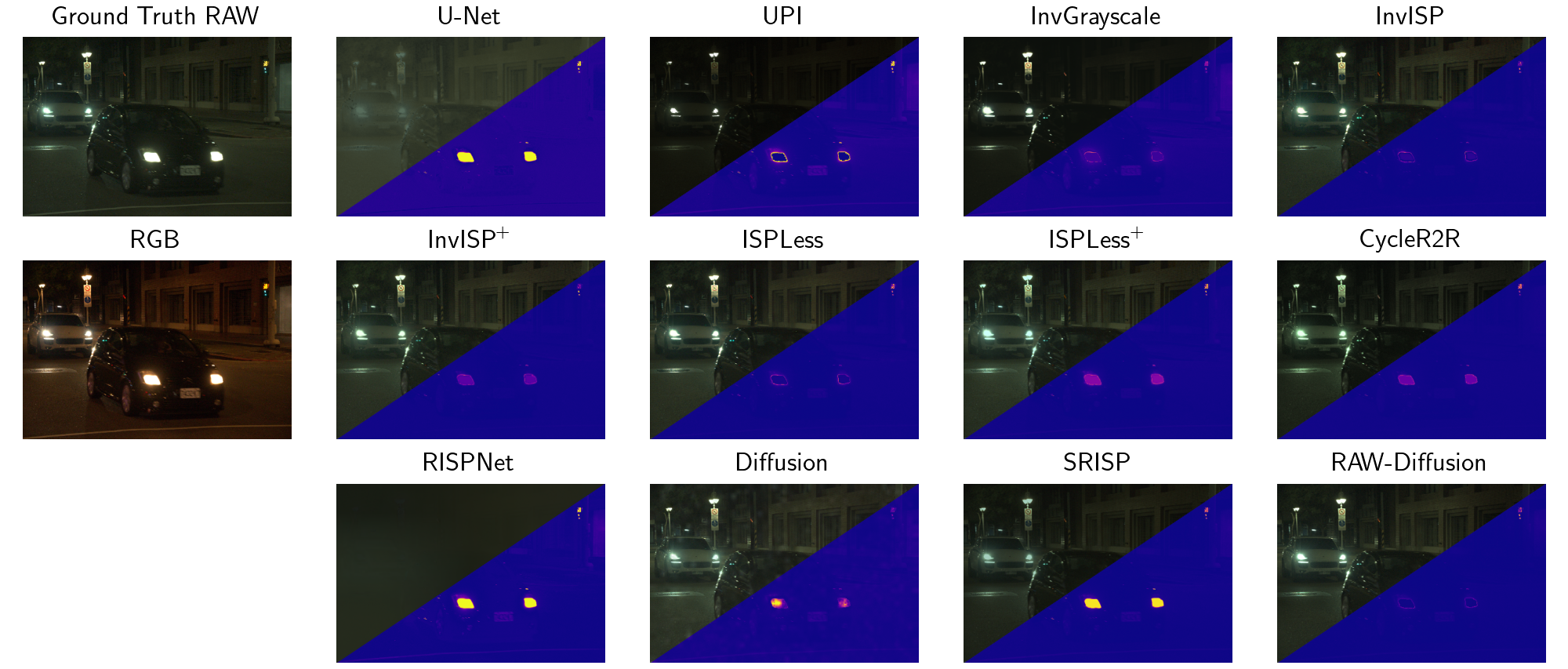}
    \caption{Qualitative results on FiveK (top) and NOD (bottom). The reconstructed RAW image and the error map are presented for each method. The RAW images are shown with a gamma correction of $1/2.2$ for visualization.}
    \vspace{-0.5cm}
    \label{fig_rgb2raw_qualitative_results}
\end{figure}

The results on the FiveK and NOD test sets for both cameras are shown in \cref{tab:exp_results_rgb2raw}.
On FiveK Nikon, U-Net records a PSNR of $26.00$, while UPI reaches a PSNR of $27.66$.
ISPLess and ISPLess$^+$ achieve a PSNR of $26.82$ and $29.83$, respectively. RAW-Diffusion attains the highest PSNR of $30.05$.
On NOD, the variability in results widens due to the challenging conditions of the dataset by the skewed distribution of the RAW values and focus on low-light scenarios across both cameras.
CycleR2R demonstrates the lowest performance with $24.93$ PSNR on NOD Nikon. InvISP$^+$ achieves a PSNR of $42.57$ and RAW-Diffusion reaches a PSNR of $44.93$.
Overall, RAW-Diffusion is able to achieve the best results across all datasets.
Qualitative results for each method are shown in Fig.~\ref{fig_rgb2raw_qualitative_results}.

\subsection{Small Data RGB2RAW Results}
\newcommand{\p}[1]{\hspace*{1.2mm}#1\hspace*{1.2mm}}

\begin{table}[tb]
  \caption{Results of training RAW-Diffusion with limited training data. Even when trained with only $25$ images, RAW-Diffusion achieves the same performance as trained on the full dataset.\vspace{-0.1cm}}
  \label{tab:exp_results_rgb2raw_small_data}
  \scriptsize
  \setlength\tabcolsep{1.5pt}
  \centering
  \begin{tabular}{r|
        cc|
        cc|
        cc|
        cc}
    \toprule
    \multirow{2}{*}{\shortstack{Training\\Images}} & \multicolumn{2}{c}{\textbf{FiveK Nikon}} & \multicolumn{2}{c}{\textbf{FiveK Canon}} & \multicolumn{2}{c}{\textbf{NOD Nikon}} & \multicolumn{2}{c}{\textbf{NOD Sony}} \\
    &  \p{PSNR} & \p{SSIM} & \p{PSNR} & \p{SSIM} & \p{PSNR} & \p{SSIM} & \p{PSNR} & \p{SSIM}  \\
    \midrule
Full & 30.05 & 0.857 & 34.01 & 0.926& 44.93 & 0.983 & 39.17 & 0.932 \\ 
\midrule
500 & 30.39 & 0.858 & 34.41 & 0.930 & 44.70 & 0.983 & 39.28 & 0.931 \\
250 & 30.50 & 0.861 & 35.93 & 0.949 & 45.11 & 0.983 & 39.22 & 0.933 \\
100 & 30.80 & 0.853 & 36.14 & 0.948 & 45.28 & 0.983 & 39.27 & 0.932 \\
50 & 30.80 & 0.837 & 35.32 & 0.928 & 45.38 & 0.982 & 39.26 & 0.931 \\
25 & 31.29 & 0.850 & 34.17 & 0.914 & 45.51 & 0.982 & 39.05 & 0.929 \\
10 & 30.54 & 0.830 & 33.44 & 0.893 & 45.31 & 0.979 & 38.66 & 0.922 \\

  \bottomrule
  \end{tabular}
\end{table}

Collecting large amounts of RAW images for each RAW camera model is extremely expensive. Therefore, we analyze the performance of RAW-Diffusion when trained with limited data by subsampling the original training set. In the following experiment, we conduct experiments with 500, 250, 100, 50, 25, and 10 training samples for training the diffusion model. 
To maintain a consistent number of training steps, the small training sets are repeated accordingly. 
In \cref{tab:exp_results_rgb2raw_small_data}, the average results on the test set of FiveK and NOD are shown.
With only $25$ training samples, RAW-Diffusion achieves a PSNR of $31.29$ and an SSIM of $0.850$ on FiveK Nikon compared to $30.05$ and $0.857$, respectively, on the full training set. Similarly, the performance remains robust across the other three DSLR datasets, even with the limited sample size.
Interestingly, the performance initially slightly improves with smaller portions of the training dataset compared to the full dataset. There are two reasons. First, the training on the full dataset is performed epoch-based while training on the subsampled datasets is iteration-based which improves performance. Additionally, the issue arises from distribution shifts between training and test sets, especially for small datasets. When increasing the number of training runs, the results better align with the expectations. The remaining variance falls within stochastic uncertainty. %
Overall, performance with limited data is very robust.
Training the diffusion model with only $10$ samples still yields a consistent performance on FiveK Nikon and NOD Nikon, achieving a PSNR of $30.54$ and $45.31$. On FiveK Canon and NOD Sony, the PSNR decreases to $33.44$ ($-0.57$) and $38.66$ ($-0.51$), respectively.
The experiment reveals that RAW-Diffusion is capable of learning a precise reconstruction with very few training examples.

\subsection{Object Detection Results}
The goal is to enable the training of neural networks for new camera sensors with very few training examples.
Unlike typical labor-intensive data labeling, we generate realistic, sensor-specific RAW images from existing large-scale RGB datasets for training.
We evaluate the object recognition performance on NOD (both Nikon and Sony cameras), using $100$ NOD images to assess adaptation to new sensors with limited target domain data.

We analyze the integration of Cityscapes \cite{cordtsCityscapesDatasetSemantic2016} and BDD100K \cite{Yu2018BDD100KAD} as large-scale RGB datasets with a large number of labeled images, and generate Cityscapes-RAW and BDD100K-RAW with RSISP~\cite{otsukaSelfSupervisedReversedImage2023}, a recent, state-of-the-art RGB2RAW conversion method, and RAW-Diffusion, each for both cameras. 
Both methods are trained with a limited number of samples from NOD. 
For RAW-Diffusion, we utilize a sampling process with 24 or 6 sampling steps via DDIM, respectively. 
The generated datasets are combined with the small original sample collection from NOD by defining a probability $p_{\text{gen}}$. 
This probability determines the likelihood of selecting a sample from the generated dataset or, from NOD, thereby ensuring the overall number of training steps remains the same.

\begin{table}[tb]
  \caption{Object detection results using Faster R-CNN evaluating the performance on 100 NOD training samples (RGB and RAW) and the integration of Cityscapes-RAW and BDD100K-RAW generated by SRISP and RAW-Diffusion.
  \vspace{-0.1cm}}
  \label{tab:exp_results_object_detection_fasterrcnn}
  \scriptsize
  \setlength\tabcolsep{1.5pt}
  \centering
  \begin{tabular}{l|
        ccc|
        ccc}
    \toprule
    & \multicolumn{3}{c}{\textbf{NOD Nikon}} & \multicolumn{3}{c}{\textbf{NOD Sony}}\\
    Training Dataset        & $\operatorname{AP}$   & $\operatorname{AP_{50}}$  & $\operatorname{AP_{75}}$ 
                            & $\operatorname{AP}$   & $\operatorname{AP_{50}}$  & $\operatorname{AP_{75}}$ \\
    \midrule
RGB     & 19.1\std{0.2} & 36.8\std{0.4} & 17.7\std{0.6}     
        & 19.2\std{0.5} & 38.2\std{0.8} & 17.6\std{0.5} \\
RAW     & 18.2\std{0.2} & 35.3\std{0.2} & 17.1\std{0.2} 
        & 18.0\std{0.1} & 35.8\std{0.3} & 16.3\std{0.4} \\
RAW + CS-RAW (SRISP)    & 23.0\std{0.6} & 43.9\std{0.4} & 21.9\std{1.5} 
                                & 23.1\std{0.4} & 46.9\std{0.4} & 20.0\std{0.1}\\
RAW + BDD (SRISP)       & 24.2\std{0.3} & 45.7\std{0.4} & 22.6\std{0.7} 
                                & 26.0\std{0.2} & 50.5\std{0.6} & 24.2\std{0.5}\\
RAW + CS-RAW (ours)     & \underline{24.7}\std{0.3}     & \underline{46.3}\std{0.6}     & \underline{23.9}\std{0.2} 
                                & \underline{26.2}\std{0.3}     & \underline{50.6}\std{0.6}     & \underline{24.9}\std{0.4}\\
RAW + BDD-RAW (ours)        & \textbf{26.5}\std{0.3}      & \textbf{49.3}\std{0.5}      & \textbf{25.3}\std{0.5} 
                                & \textbf{28.6}\std{0.1}      & \textbf{55.2}\std{0.4}      & \textbf{26.4}\std{0.3}\\ 
  
  \bottomrule
  \end{tabular}
\end{table}

\begin{table}[tb]
  \caption{Performance on the test set using YOLOv8 for training on NOD and combinations of NOD with generated datasets. 
  \vspace{-0.1cm}}
  \label{tab:exp_results_object_detection_yolo}
  \scriptsize
  \setlength\tabcolsep{1.5pt}
  \centering
  \begin{tabular}{l|
        ccc|
        ccc}
    \toprule
    & \multicolumn{3}{c}{\textbf{NOD Nikon}} & \multicolumn{3}{c}{\textbf{NOD Sony}}\\
    Training Dataset        & $\operatorname{AP}$   & $\operatorname{AP_{50}}$  & $\operatorname{AP_{75}}$ 
                            & $\operatorname{AP}$   & $\operatorname{AP_{50}}$  & $\operatorname{AP_{75}}$ \\
    \midrule
RGB         & 22.7\std{1.7} & 32.9\std{10} & 22.5\std{1.8} 
            & 18.4\std{0.3} & 33.8\std{0.2} & 18.4\std{0.4} \\
RAW         & 25.8\std{0.5}	& 45.5\std{0.8}	& 25.7\std{0.9} 
            & 27.6\std{0.4}	& 49.3\std{0.2}	& 27.0\std{0.9} \\
RAW + CS-RAW (SRISP)    & 29.1\std{0.1}	& 48.6\std{0.2}	& 29.4\std{0.1} 
                                & 29.5\std{0.4}	& 51.7\std{0.6}	& 29.1\std{0.7} \\
RAW + BDD-RAW (SRISP)       & \underline{32.0}\std{0.3}	& \underline{53.1}\std{0.5}	& \underline{32.1}\std{0.4} 
                                & \underline{32.0}\std{0.7}	& \underline{55.3}\std{0.8}	& \underline{31.6}\std{0.8} \\
RAW + CS-RAW (ours)     & 29.9\std{0.3}	& 50.2\std{0.5}	& 30.7\std{0.7}
                                & 31.0\std{0.3}	& 54.8\std{0.5}	& 31.0\std{0.5} \\
RAW + BDD-RAW (ours)        & \textbf{32.6}\std{0.1}	& \textbf{54.3}\std{0.2}	& \textbf{32.6}\std{0.7} 
                                & \textbf{33.6}\std{0.1}	& \textbf{58.3}\std{0.2}	& \textbf{33.8}\std{0.0} \\
  \bottomrule
  \end{tabular}
\end{table}

The results on the test set of NOD Nikon and Sony using Faster R-CNN, a two-stage method, are shown in \cref{tab:exp_results_object_detection_fasterrcnn}.
Training on the original RGB and RAW images reaches an AP of $19.1$ and $18.2$, respectively, on NOD Nikon. Incorporating  Cityscapes-RAW and BDD100K-RAW, generated by SRISP, enhances the AP to $23.0$ and $24.2$ AP. The training on the generated datasets by RAW-Diffusion raises the performance to $24.7$ and $26.5$.
Additionally, we perform the evaluation using YOLOv8, a one-stage method. The results in \cref{tab:exp_results_object_detection_yolo} show a similar outcome, where models trained on both Cityscapes-RAW and BDD100K-RAW datasets generated by our method outperform models trained on the same datasets converted by SRISP. These results highlight the importance of generating high-fidelity RAW images when converting RGB datasets to a target domain, for increasing performance on a downstream task such as object detection.

\begin{table}[tb]
  \caption{Zero-shot object detection results using Faster R-CNN. The models are trained exclusively on the generated datasets and evaluated on the NOD test set.\vspace{-0.1cm}}
  \label{tab:exp_results_object_detection_fasterrcnn_zeroshot}
  \scriptsize
  \setlength\tabcolsep{1pt}
  \centering
  \begin{tabular}{l|
        ccc|
        ccc}
    \toprule
    & \multicolumn{3}{c}{\textbf{NOD Nikon}} & \multicolumn{3}{c}{\textbf{NOD Sony}} \\
    Training Dataset 
        & $\operatorname{AP}$ 
        & $\operatorname{AP_{50}}$ 
        & $\operatorname{AP_{75}}$ 
        & $\operatorname{AP}$ 
        & $\operatorname{AP_{50}}$ 
        & $\operatorname{AP_{75}}$ \\
    \midrule
CS-RAW (SRISP) 
        & 7.7\std{1.0} & 16.8\std{2.7} & 6.4\std{0.6} 
        & 3.8\std{1.1} & 9.3\std{2.8} & 2.6\std{0.5}\\
BDD-RAW (SRISP)  
        & \underline{18.4}\std{0.3} & \underline{35.9}\std{0.3} & \underline{15.8}\std{0.5} 
        & \underline{16.9}\std{0.6} & \underline{34.2}\std{0.6} & \underline{14.5}\std{0.8} \\
CS-RAW (ours) 
        & 12.0\std{0.7} & 23.4\std{1.6} & 11.4\std{0.5} 
        & 13.7\std{1.3} & 29.4\std{2.3} & 11.7\std{0.8} \\
BDD-RAW (ours)          
        & \textbf{22.0}\std{0.1} & \textbf{43.1}\std{0.3} & \textbf{18.8}\std{0.5} 
        & \textbf{21.6}\std{0.1} & \textbf{43.7}\std{0.4} & \textbf{18.7}\std{0.2}\\
  \bottomrule
  \end{tabular}
\end{table}

In the next experiment, we assess zero-shot performance when training on the generated datasets exclusively (see \cref{tab:exp_results_object_detection_fasterrcnn_zeroshot}). 
While Cityscapes-RAW generated by SRISP achieves an AP of $7.7$ and RAW-Diffusion reaches an AP of $12.0$, the results indicate that the domain shift from Cityscapes, notably its limitation to daylight images, is too significant for effective training without any NOD images. 
However, the larger dataset size and greater diversity in BDD100K(-RAW) increase performance, achieving $18.4$ for SRISP and $22.0$ for RAW-Diffusion.
The zero-shot results using YOLO are shown in the supplementary material.

\begin{figure}[t]
    \centering
    \includegraphics[width=\columnwidth]{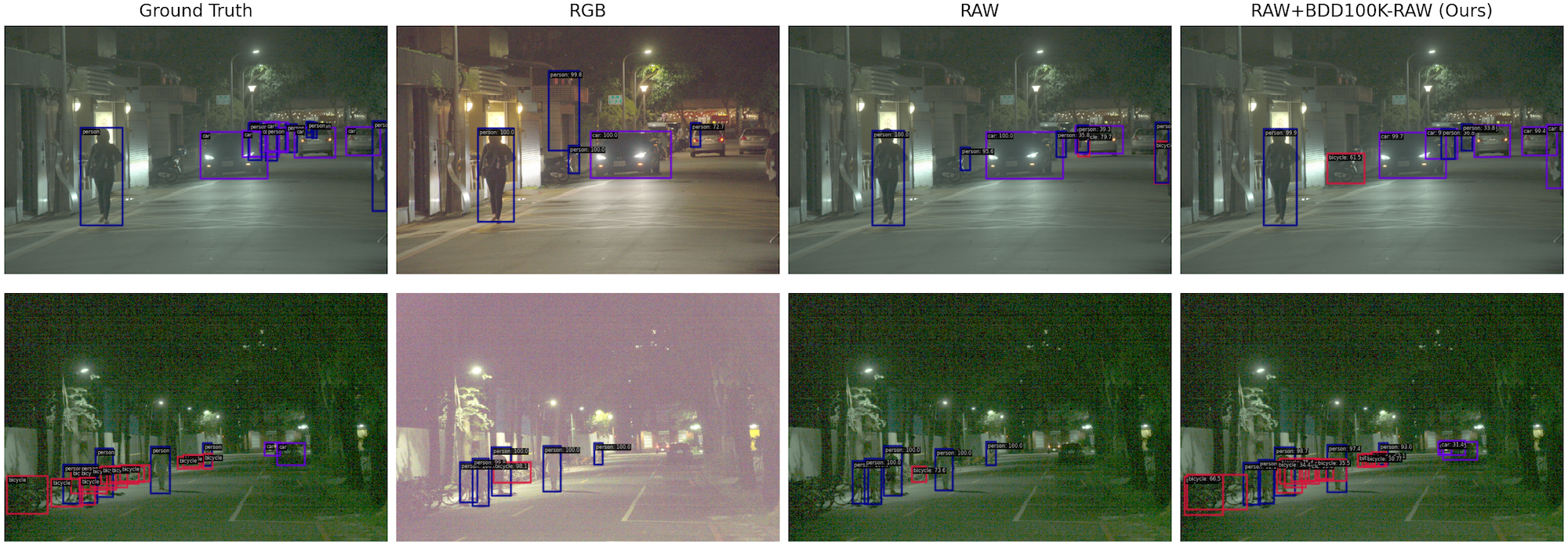}
    \caption{Qualitative results on NOD Nikon (top) and Sony (bottom) for training on the RGB and RAW dataset and a combination with BDD100K-RAW generated by RAW-Diffusion. Please refer to the supplementary material for further qualitative analysis.}
    \label{fig_od_qualitative_results}
\end{figure}

\subsection{Analyses and Ablation Studies}
\label{sec_experiments_analysis}
We conduct several analyses and ablation studies to examine the fundamental components of the proposed method and assess their impact on the performance.

\vspace{-0.2cm}
\paragraph{RAW-Diffusion Architecture.}

First, we analyze the architecture of the diffusion model. The results in \cref{tab:exp_analysis_diffusion} reveal that predicting the noise, which is a common parametrization in the image domain, does not translate effectively to RAW images, leading to $5.55$ PSNR on NOD Nikon. Instead, we achieve significantly better results by directly learning the reconstruction of the RAW image.
Furthermore, we explore the effect of the RGB-guidance module by alternatively concatenating the noisy RAW image with the RGB image, and examine the role of the hyperbolic tangent function as the final activation.
The results show that the modifications reduce the PSNR from $44.93$ to $43.45$ and $39.74$, respectively.

\begin{table}[b]
    \centering
    \begin{minipage}[t]{\columnwidth}
        \centering
        \caption{Impact of the RAW-Diffusion architecture. Our model directly predicts the RAW image, incorporates an RGB-guidance module for conditioning, and a hyperbolic tangent activation.\vspace{-0.2cm}}
        \label{tab:exp_analysis_diffusion}
        \scriptsize
        \centering
        \begin{tabular}{l|
            cc|
            cc}
        \toprule
        & \multicolumn{2}{c}{\textbf{NOD Nikon}} & \multicolumn{2}{c}{\textbf{NOD Sony}} \\
        \textbf{Experiment} & PSNR & SSIM & PSNR & SSIM  \\
        \midrule
        RAW-Diffusion           & 44.93     & 0.983     & 39.17     & 0.932 \\
        \quad predict noise  & 5.55      & 0.001     & 5.04      & 0.001 \\
        \quad concat RGB        & 43.45     & 0.978     & 38.15     & 0.920 \\
        \quad without TanH      & 39.74     & 0.934     & 36.93     & 0.850 \\
        \bottomrule
    \end{tabular}
    \end{minipage}\hfill\\
    \vspace{0.5cm}
    \begin{minipage}[t]{\columnwidth}
        \centering
        \caption{Analysis of the RAW reconstruction performance with a reduced number of sampling steps via DDIM. RAW-Diffusion achieves strong performance with very few sampling steps.\vspace{-0.2cm}}
        \label{tab:exp_analysis_ddim}
        \scriptsize
        \begin{tabular}{lr|
            cc|
            cc}
        \toprule
        & & \multicolumn{2}{c}{\textbf{NOD Nikon}} & \multicolumn{2}{c}{\textbf{NOD Sony}} \\
        & \textbf{Timesteps} & PSNR & SSIM & PSNR & SSIM  \\
        \midrule
        \textbf{DDPM}   & 1000  & 44.97 & 0.982 & 38.79 & 0.931 \\
        \midrule
        \multirow{8}{*}{\textbf{DDIM}} 
                        & 1000  & 45.04 & 0.983 & 39.02 & 0.931 \\
                        & 100   & 45.05 & 0.983 & 39.07 & 0.931 \\
                        & 48    & 45.05 & 0.983 & 39.12 & 0.932 \\
                        & 24    & 44.91 & 0.983 & 39.16 & 0.932 \\
                        & 12    & 45.11 & 0.983 & 39.28 & 0.932 \\
                        & 6     & 45.16 & 0.984 & 39.29 & 0.931 \\
                        & 3     & 44.92 & 0.980 & 38.75 & 0.926 \\
                        & 2     & 44.19 & 0.975 & 36.27 & 0.902 \\
      \bottomrule
      \end{tabular}
    \end{minipage}
\end{table}

\vspace{-0.2cm}
\paragraph{Sampling Steps.}

In the next experiment, we examine DDIM \cite{songDenoisingDiffusionImplicit2021} for accelerating the sampling process. While DDPM performs $T$ iterations, DDIM reduces the number of sampling steps via a non-Markovian diffusion process. The results in \cref{tab:exp_analysis_ddim} show that RAW-Diffusion achieves consistent performance with as few as $6$ DDIM sampling steps compared to $1000$ step DDPM.
Reducing the steps to $3$ and $2$ drops PSNR values for NOD Sony from $39.29$ to $38.75$ and $36.27$, respectively.

\vspace{-0.2cm}
\paragraph{Loss Functions.}

The diffusion model is trained with a combination of mean squared error ($\lossmse$), L1 ($\lossLone$), and logL1 loss ($\losslogLone$) to learn the reconstruction of the RAW images. 
In the analysis summarized in \cref{tab:exp_analysis_losses}, we study the impact of the individual component by progressively adding the losses. 
On NOD Nikon, the reconstruction quality consistently improves as we add more diverse loss terms.
On NOD Sony, the performance across all variants stays comparable, albeit with a minor decrease when training with only $\lossmse$ and $\lossLone$.

\begin{table}[t]
    \centering
    \begin{minipage}{0.48\textwidth}
        \centering
        \caption{Impact of loss components. The combination of $\lossmse$, $\lossLone$, and $\losslogLone$ increases the precision.\vspace{-0.15cm}}
        \label{tab:exp_analysis_losses}
        \scriptsize
        \begin{tabular}{
            >{\centering\arraybackslash}p{0.65cm}
            >{\centering\arraybackslash}p{0.6cm}
            >{\centering\arraybackslash}p{0.75cm}
            |
            cc|
            cc}
            \toprule
            & & & \multicolumn{2}{c}{\textbf{NOD Nikon}} & \multicolumn{2}{c}{\textbf{NOD Sony}} \\
            $\lossmse$ & $\lossLone$ & $\losslogLone$ & PSNR & SSIM & PSNR & SSIM  \\
            \midrule
            \checkmark & & &  43.10 & 0.970 & 39.03 & 0.936  \\
            \checkmark & \checkmark & &  44.35 & 0.980 & 38.85 & 0.926 \\
            \checkmark & \checkmark & \checkmark &  44.93 & 0.983 & 39.17 & 0.932 \\
            \bottomrule
        \end{tabular}
    \end{minipage}\hfill\\
    \vspace{0.5cm}
    \begin{minipage}{0.47\textwidth}
        \centering
        \caption{Analysis of different RGB-Guidance backbones. RAW-Diffusion works well with different backbones.\vspace{-0.15cm}}
        \label{tab:exp_analysis_rgb_backbone}
        \scriptsize
        \begin{tabular}{l|c|
            cc|
            cc}
            \toprule
            \multicolumn{2}{c}{} &\multicolumn{2}{c}{\textbf{NOD Nikon}} & \multicolumn{2}{c}{\textbf{NOD Sony}} \\
            \textbf{Backbone} & Size & PSNR & SSIM & PSNR & SSIM  \\
            \midrule
            EDSR    & 25M    & 44.93 & 0.983 & 39.17 & 0.932 \\
            RRDB    & 27M    & 45.18 & 0.984 & 38.95 & 0.930 \\
            NAFNet  & 81M    & 44.70 & 0.982 & 39.11 & 0.926 \\
            \bottomrule
        \end{tabular}
    \end{minipage}
\end{table}

\vspace{-0.2cm}
\paragraph{RGB-Guidance Backbones.}

We conduct an analysis of different backbones for the RGB-guidance module, for which, the standard EDSR-based backbone~\cite{limEnhancedDeepResidual2017} is substituted with RRDB~\cite{wangESRGANEnhancedSuperresolution2019} and NAFNet~\cite{chenSimpleBaselinesImage2022} backbones by incorporating the respective feature extraction modules. 
The results in \cref{tab:exp_analysis_rgb_backbone} demonstrate that the RGB-guidance module achieves robust performance across different backbones. 
Notably, the most extensive backbone, NAFNet, with a total of 81M parameters, does not enhance the precision further.

\paragraph{RGB-Guidance Conditioning.}

Next, we provide a comparison of the different conditioning architectures, namely ControlNet~\cite{zhangAddingConditionalControl2023}, T2I-Adapter~\cite{mouT2IAdapterLearningAdapters2024}, StableSR~\cite{wangExploitingDiffusionPrior2024}, and RAW-Diffusion.
\textit{ControlNet} uses a simple addition after the first projection layer without sophisticated injection mechanism.
\textit{T2I-Adapter} uses a downsampling encoder (which performed worse in our tests) and adds features only within the encoder.
\textit{StableSR} uses a time-aware embedding encoder with downsampling (our time-independent encoder is more efficient). 
The injection is different, modulating only the features before the residual connection, showing reduced performance in our preliminary tests.
The experiments are performed using the same training pipeline. 
The results on NOD Nikon are shown in \cref{tab:exp_analysis_guidance_architectures}, demonstrating the superior performance of RAW-Diffusion.

\begin{table}[]
  \scriptsize
  \caption{Analysis of different conditioning architectures\vspace{-0.15cm}}
  \label{tab:exp_analysis_guidance_architectures}
  \centering
  \begin{tabular}{
        lcccc}
   \toprule
     & \multicolumn{1}{c}{ControlNet} & \multicolumn{1}{c}{T2I-Adapter} & \multicolumn{1}{c}{StableSR} & \multicolumn{1}{c}{RAW-Diffusion}\\
    \midrule
 PSNR & 42.15\std{0.14} & 42.13\std{0.28}& 44.43\std{0.18}	 & \textbf{44.93}\stdt{0.15} \\
 SSIM & 0.973\std{0.00} & 0.969\std{0.00}   & \textbf{0.983}\std{0.00}	 & \textbf{0.983}\stdt{0.00} \\
  \bottomrule
  \end{tabular}
  \vspace{-0.2cm}
\end{table}

Besides distinctions in conditioning, RAW-Diffusion introduces further general differences, such as loss functions, output activation, RAW-prediction, and a specialized encoder, all essential for detailed RAW image generation.

\vspace{-0.2cm}
\paragraph{Dataset Mixing Ratio.}
 
\begin{figure}[t]
    \centering
    \includegraphics[width=\columnwidth]{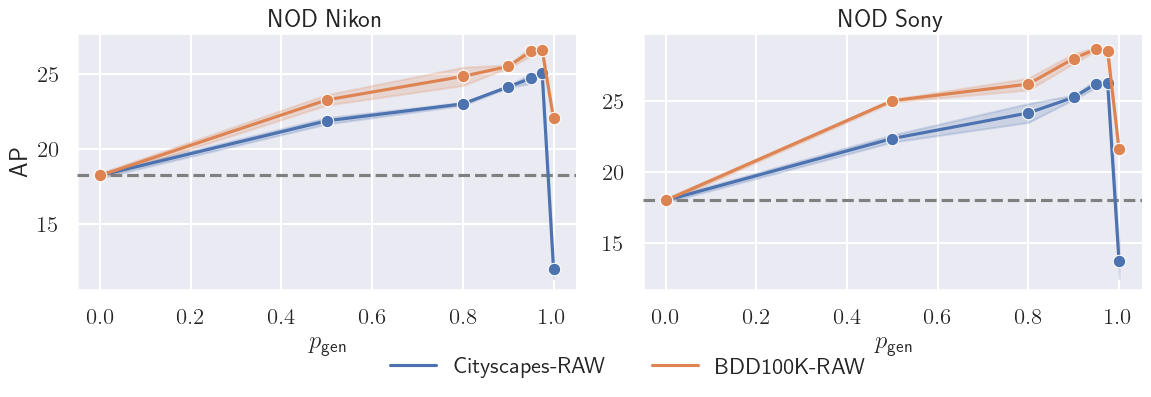}
    \vspace{-0.7cm}
    \caption{Analysis of different probabilities $p_{\text{gen}}$.
    The performance improves with increasing probability of sampling from Cityscapes-RAW and BDD100K-RAW, respectively, reaching a peak before training exclusively on the generated datasets.}
    \label{fig_od_analysis_f}
\end{figure}

Finally, we study the contribution of Cityscapes-RAW and BDD100K-RAW generated by RAW-Diffusion with different mixing ratios. The probability $p_{\text{gen}}$ defines the likelihood of sampling a data sample from the generated dataset. 
The mean performance and standard deviation on NOD Nikon and Sony are shown in Fig. \ref{fig_od_analysis_f}.
The results of training exclusively on the original images, i.e., $p_{\text{gen}} = 0$, are highlighted by a dashed line. 
The experiment reveals that best performance is achieved by training primarily on Cityscapes-RAW and BDD100K-RAW, respectively, complemented with a minor portion of the original image, i.e., $p_{\text{gen}}$ of $95\%$ or $97.5\%$.
When $p_{\text{gen}} = 1$, the training is performed on the generated datasets exclusively without any NOD images.

\section{Limitations and Future Work}

Although our method has proven to be data-efficient, dedicated training for each camera sensor is required. 
We plan to work on the generalization ability of our method via conditioning for a multi-sensor model.
We also emphasize that as a generative model, RAW-Diffusion cannot mitigate any existing biases in the datasets. Metadata is missing for a bias/fairness evaluation.
However, we would like to target the removal of dataset bias in our future work. 
In object detection, images from NOD with overlapping or small objects reveal the limitations of standard models, suggesting the adaptation of specialized models \cite{zhengProgressiveEndtoendObject2022,yuanSmallObjectDetection2023}.

\section{Conclusion}

In this work, we presented RAW-Diffusion, a novel diffusion-based method for generating high-fidelity RAW images guided by RGB images.
RAW-Diffusion introduces an RGB-guidance module for extracting low-level and high-level guidance features.
These features are incorporated into the iterative denoising process via RGB-guided residual blocks to modulate the diffusion features.
RAW-Diffusion directly predicts the RAW images and learns the precise reconstruction through iterative denoising.
Comprehensive experiments on four DSLR camera datasets show the superior performance of RAW-Diffusion over current state-of-the-art methods. 
Furthermore, we demonstrate that RAW-Diffusion exhibits remarkable effectiveness, even when trained on a limited dataset with $25$ images or fewer.
On the downstream task, RAW-Diffusion enables the generation of realistic RAW datasets, like Cityscapes-RAW and BDD100K-RAW, from large-scale RGB datasets, facilitating the training on new camera sensors with very few training samples from the target domain.

\clearpage

\bibliographystyle{ieee_fullname}
\bibliography{bibliography}

\clearpage

\setcounter{section}{0}
\renewcommand{\thesection}{\Alph{section}}

\twocolumn[
  \begin{center}
    \Large \bf RAW-Diffusion: RGB-Guided Diffusion Models \\for High-Fidelity RAW Image Generation \\(Supplementary Material)
  \end{center}
]

In the supplementary material, the RAW-Diffusion hyper-parameters and the training details for object detection with Faster R-CNN and YOLOv8 are described. 
Furthermore, we provide additional quantitative and qualitative results for the RGB2RAW reconstruction and downstream object detection experiments.

\section{Hyper-parameters}
The hyper-parameters for the training, architecture, and diffusion process of RAW-Diffusion are detailed in \cref{sm_tab_hyperparameters}.
The number of base features is denoted by $N_{\text{Features}}$ and the number of groups used in the group normalization is denoted by $N_{\text{Norm,Groups}}$.
Further implementation details can be found in the published code: \textcolor{magenta}{\url{https://github.com/SonyResearch/RAW-Diffusion}}. %

\section{Object Detection Training Details}
The experiments are performed using a Faster R-CNN \cite{renFasterRCNNRealtime2015} and YOLOv8 \cite{yolov8}. 
Faster R-CNN has a ResNet-50 backbone pretrained on ImageNet, and it is trained with RGB and RAW images normalized using the corresponding mean and standard deviation of the dataset.
We apply random ﬂip, random resize, and cropping as data augmentation, use an image size of $416 \times 640$, and train the network for $48$ epochs.
In particular, for YOLOv8, the images are normalized to $[0,1]$, we use the same size of $416 \times 640$, and the model is finetuned from a COCO \cite{linMicrosoftCOCOCommon2014} pretrained checkpoint for 10 epochs using random flip augmentation.
In addition, the RAW images are reduced to three channels by averaging the two green channels.

For the generation of the RAW datasets (Cityscapes-RAW and BDD100K-RAW) from large-scale RGB datasets, SRISP and RAW-Diffusion are trained on the images of the object detection datasets with a limited number of samples.
When combining the original NOD images and the generated datasets, $p_{\text{gen}}$ is set to $0.95$.

\section{Qualitative RGB2RAW Results}

\begin{figure*}[]
    \centering
    \includegraphics[width=0.95\textwidth]{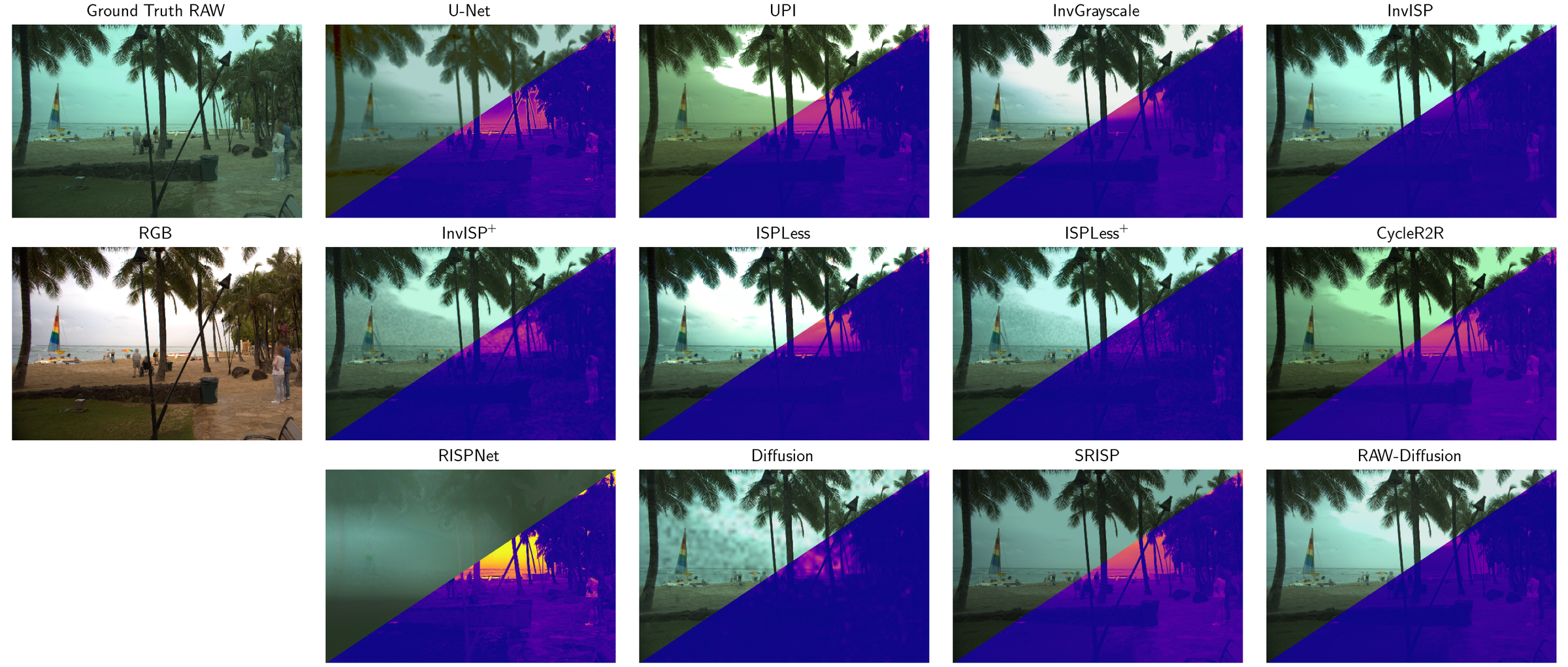}
    \includegraphics[width=0.95\textwidth]{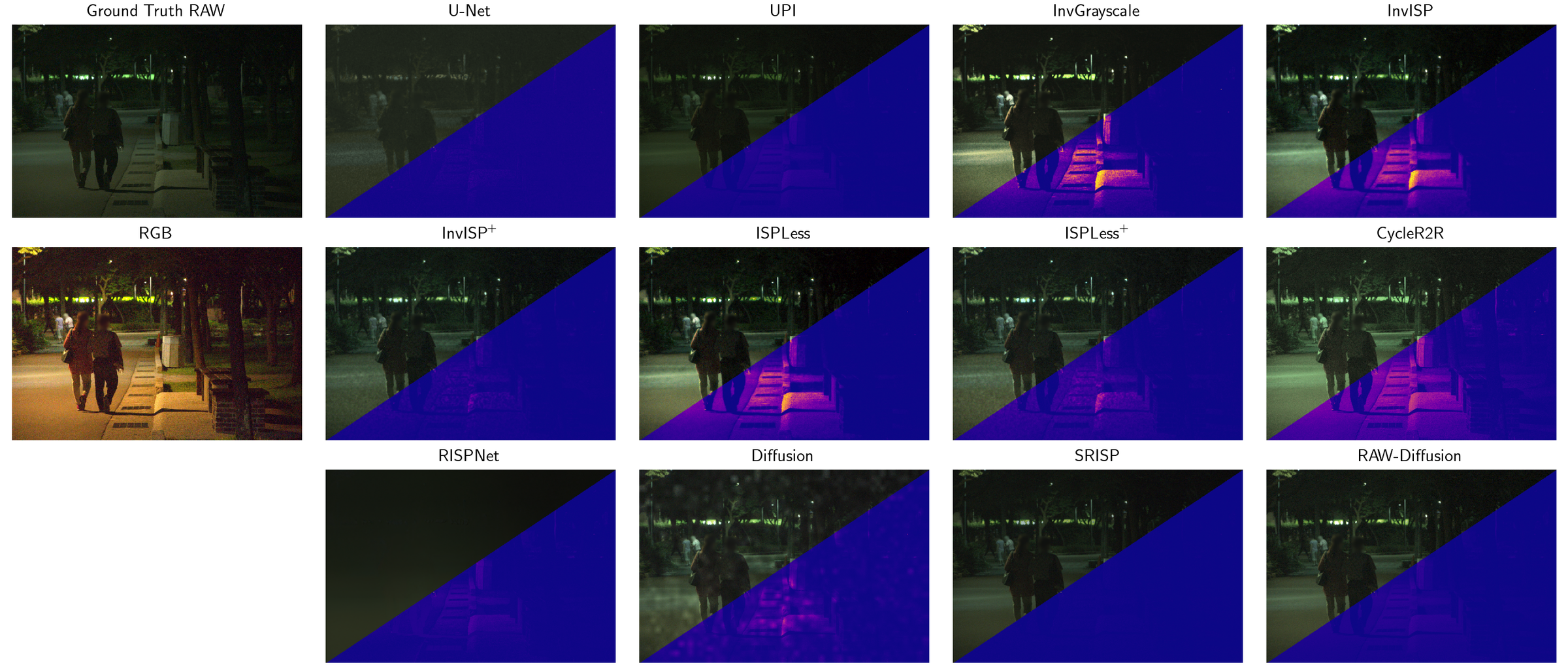}
    \includegraphics[width=0.95\textwidth]{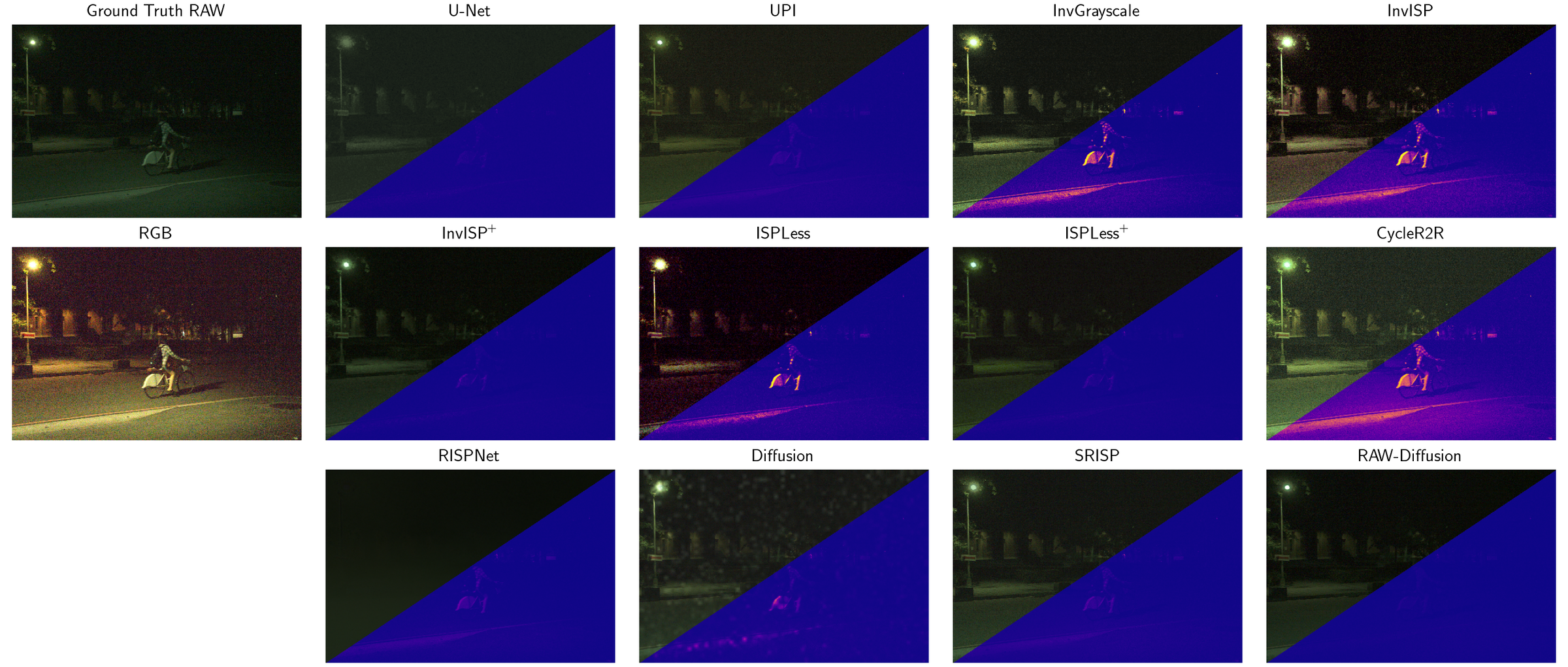}

    \caption{Qualitative results on FiveK Canon (top), NOD Nikon (middle), and NOD Sony (bottom). The reconstructed RAW image and the error map are presented for each method. The RAW images are shown with a gamma correction of $1/2.2$ for visualization.}
    \vspace{-0.5cm}
    \label{sm_fig_rgb2raw_qualitative_results}
\end{figure*}

Additional qualitative RAW reconstruction results are shown in Fig.~\ref{sm_fig_rgb2raw_qualitative_results}.
The comparison includes all methods, i.e.,
U-Net \cite{condeReversedImageSignal2022}, UPI \cite{brooksUnprocessingImagesLearned2019}, InvGrayscale \cite{xiaInvertibleGrayscale2018}, InvISP \cite{xingInvertibleImageSignal2021}, InvISP$^{+}$ \cite{xingInvertibleImageSignal2021}, ISPLess \cite{dattaEnablingISPlessLowpower2023}, ISPLess$^{+}$ \cite{dattaEnablingISPlessLowpower2023}, CycleR2R \cite{liEfficientVisualComputing2024}, RISPNet \cite{dongRISPNetNetworkReversed2023}, Diffusion \cite{liuInvertingImageSignal2022}, SRISP \cite{otsukaSelfSupervisedReversedImage2023}, and RAW-Diffusion.

\section{Detailed Object Detection Results}

In \cref{tab:exp_results_object_detection_fasterrcnn_full} and \cref{tab:exp_results_object_detection_yolo_full}, extended object detection results are presented using Faster R-CNN and YOLOv8, respectively. Performance metrics, including $\operatorname{AP}$, $\operatorname{AP_{50}}$, $\operatorname{AP_{75}}$, $\operatorname{AP_{\text{S}}}$, $\operatorname{AP_{\text{M}}}$, and $\operatorname{AP_{\text{L}}}$, are detailed. 
The experiments evaluate the adaptation performance on the target domain by training on a limited subset of $100$ sample from the original NOD, denoted by RGB and RAW, respectively, and the combination with the generated datasets by SRISP and RAW-Diffusion, i.e., Cityscapes-RAW (SRISP), BDD100K-RAW (SRISP), Cityscapes-RAW (ours), and BDD100K-RAW (ours).

Additionally, the zero-shot performance is shown in \cref{tab:exp_results_object_detection_fasterrcnn_zeroshot_full} and \cref{tab:exp_results_object_detection_yolo_zeroshot_full} using Faster R-CNN and YOLOv8, respectively. The models are trained exclusively on the generated RAW datasets and evaluated on the test set of NOD.

\section{Qualitative Object Detection Results}
\label{od_qualitative_results}

\begin{figure*}[]
    \centering
    \includegraphics[width=0.8\textwidth]{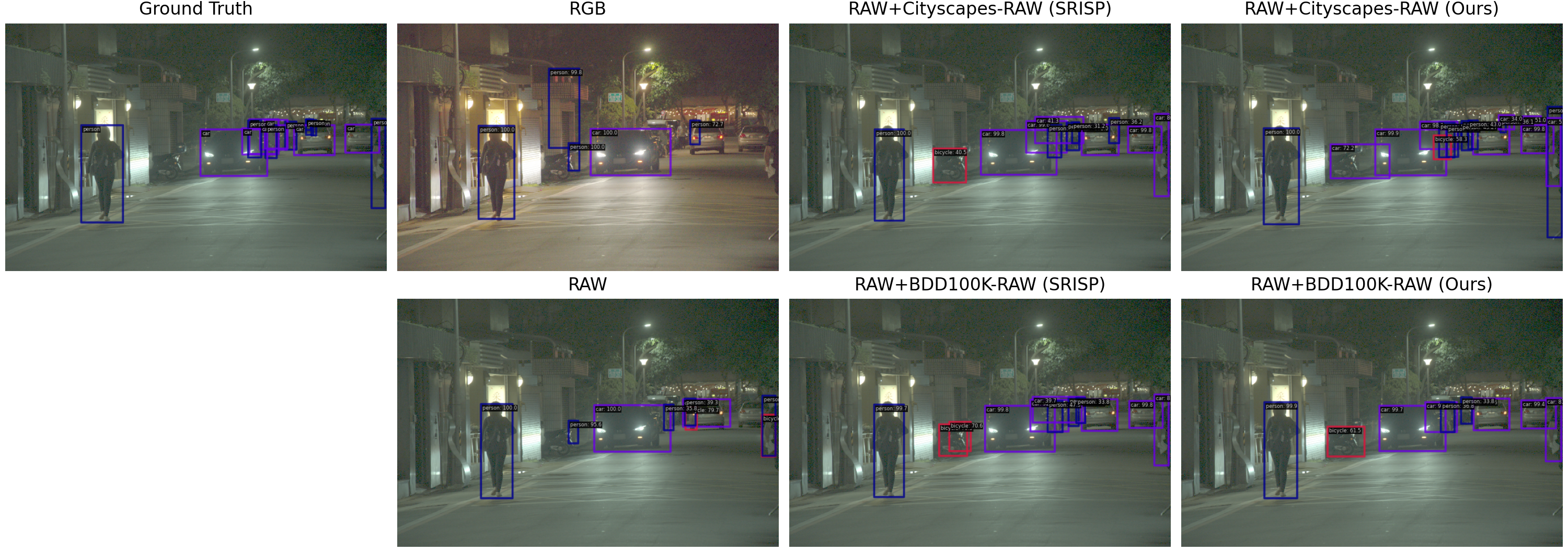}
    \includegraphics[width=0.8\textwidth]{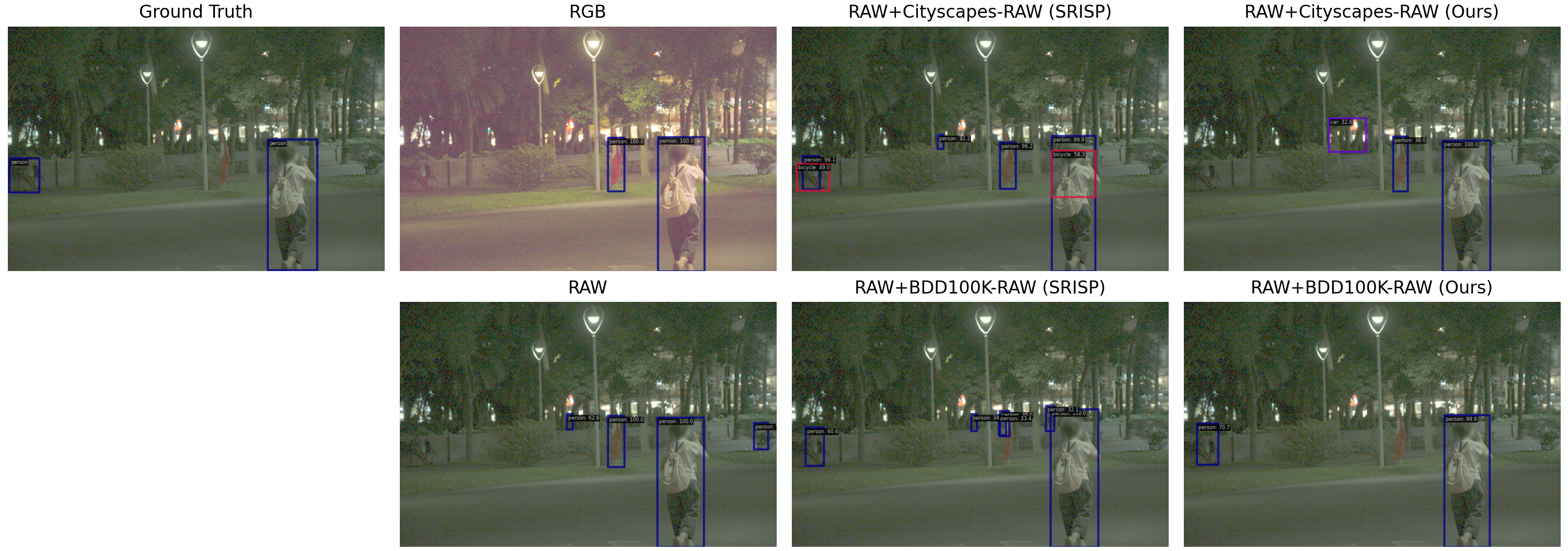}
    \includegraphics[width=0.8\textwidth]{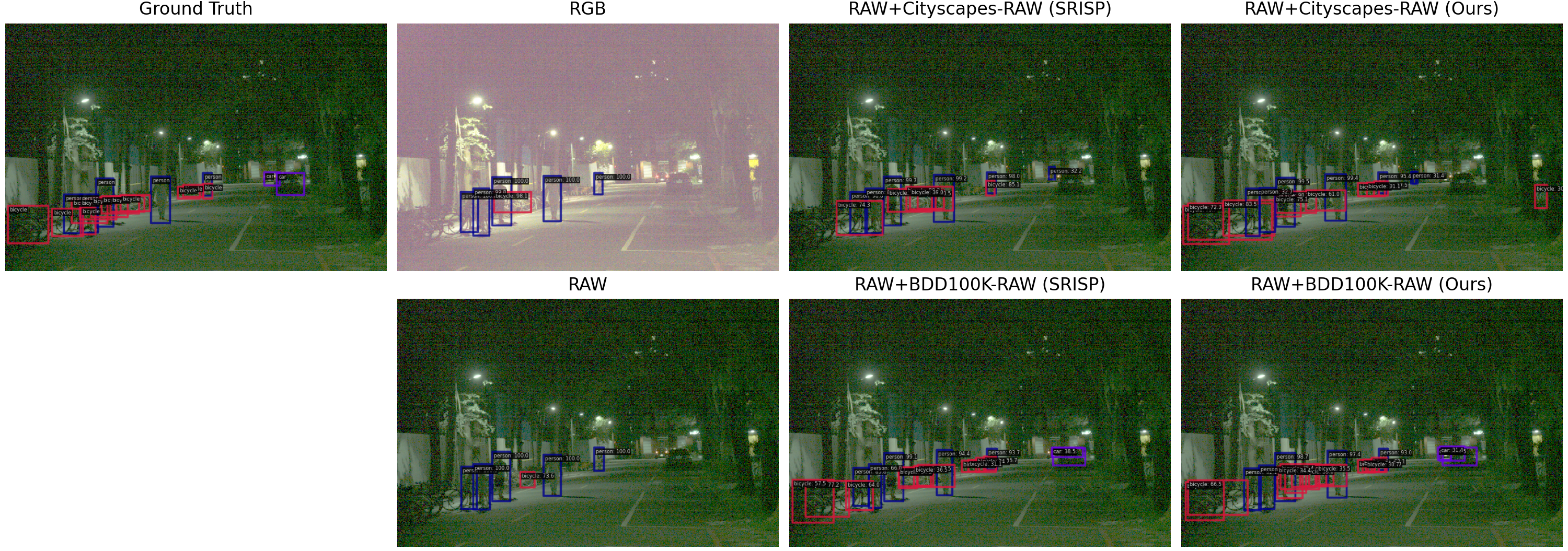}
    \includegraphics[width=0.8\textwidth]{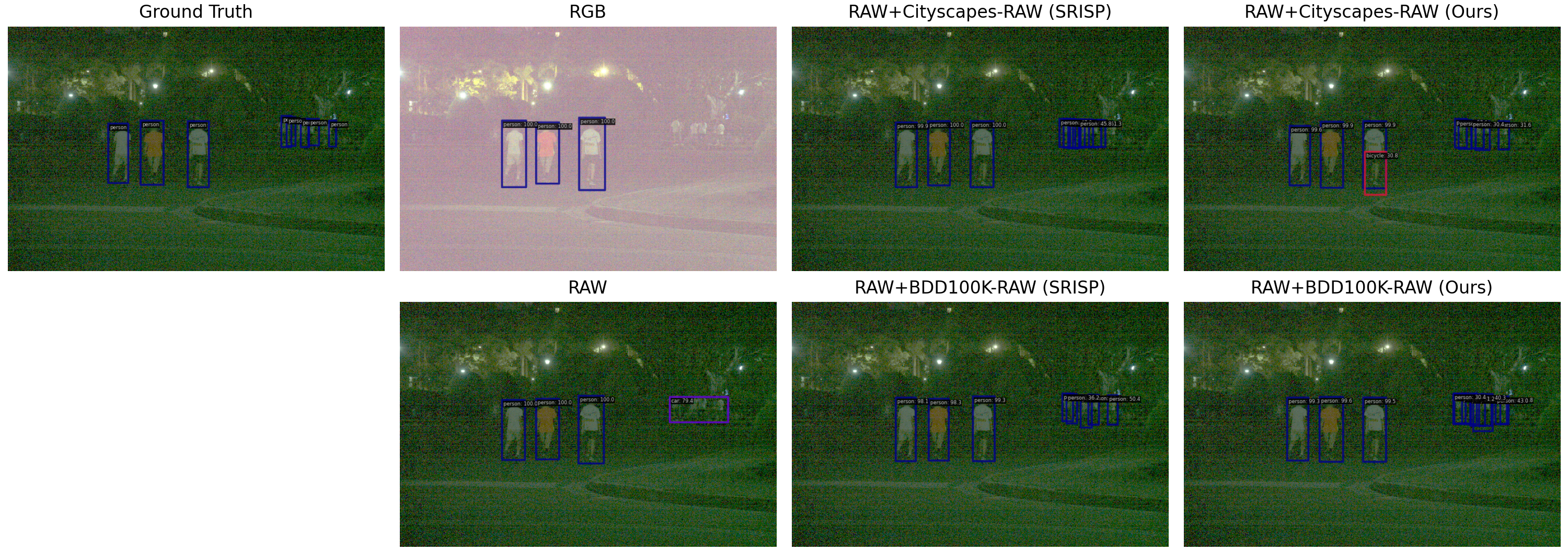}

    \caption{Qualitative object detection results from various models on the test set of NOD Nikon (first and second row) and Sony (third and forth row).}
    \vspace{-0.5cm}
    \label{sm_fig_od_qualitative_results}
\end{figure*}

In \cref{sm_fig_od_qualitative_results}, qualitative results from different object detection models using Faster R-CNN are presented. The models are trained on RGB images, RAW images, or a combination with the generated datasets. The experiment shows the advantages of RAW images compared to RGB images, especially in low-light scenarios. Furthermore, the integration of the generated RAW datasets, i.e., Cityscapes-RAW and BDD100K-RAW generated by SRISP and RAW-Diffusion, improves the precision.

\section{Training on RGB datasets for object detection on RAW images}

We analyze if object detection performance on RAW images improves by directly adding RGB samples to the original RAW training set instead of adding converted RAW images.
The results of this experiment on NOD Nikon with Cityscapes and BDD100K, respectively, are shown in \cref{tab:exp_results_od_rgb}.
Adding RGB training samples does not improve object detection performance as much as our generated RAW dataset. This highlights that the quality and distribution of the training dataset are crucial.

\begin{table*}[b]
    \centering
    \small
    \caption{Hyper-parameters of RAW-Diffusion.\vspace{-0.2cm}}

    \begin{tabular}{llc}
        \toprule
        & Hyper-parameter & Value \\
        \midrule
        \multirow{5}{*}{Training} & Training Steps & 70k \\
        & Optimizer & AdamW, $\beta$=[0.9, 0.999] \\
        & Weight Decay & 0.0 \\
        & Learning Rate & $0.0001$, linearly decreasing to zero \\
        & Batch Size & 4 \\
        \midrule
        \multirow{7}{*}{Architecture} & $N_\text{ResBlocks}$ & 2 \\
        & $N_{\text{Features}}$ & 32 \\
        & Feature Expansion & $(1, 1, 2, 2, 4, 4)$ \\
        & $N_\text{Norm,Groups}$ & 8 \\
        & Attention Block Resolutions & $16\times16$, $8\times8$ \\
        & $N_{\text{GM,ResBlocks}}$ & 4 \\ 
        & $C_{\text{GM,Features}}$ & 64 \\ 
        \midrule
\multirow{2}{*}{Diffusion} & Schedule & Linear, $\beta_1 = 0.0001$ to $\beta_T = 0.02$ \\
        & Steps & 1000 \\
        \bottomrule
    \end{tabular}
    \vspace{9cm}
    \label{sm_tab_hyperparameters}
\end{table*}

\begin{table*}[]
  \caption{Object detection results using Faster R-CNN that is trained using 100 NOD training samples (RGB and RAW). Additionally, Cityscapes-RAW and BDD100K-RAW generated by SRISP and RAW-Diffusion are integrated.
  The best result is shown in bold, and the second best underlined. \vspace{-0.1cm}
  }
  \label{tab:exp_results_object_detection_fasterrcnn_full}
  \small
  \centering
  \begin{tabular}{l|
        cccccc}
    \toprule
    & \multicolumn{6}{c}{\textbf{NOD Nikon}} \\
    Training Dataset & $\operatorname{AP}$ & $\operatorname{AP_{50}}$ & $\operatorname{AP_{75}}$ & $\operatorname{AP_{\text{S}}}$ & $\operatorname{AP_{\text{M}}}$ & $\operatorname{AP_{\text{L}}}$ \\
    \midrule
RGB     & 19.1\std{0.2} & 36.8\std{0.4} & 17.7\std{0.6} & 2.0\std{0.1} & 16.8\std{0.4} & 44.6\std{0.3}\\
RAW     & 18.2\std{0.2} & 35.3\std{0.2} & 17.1\std{0.2} & 1.6\std{0.0} & 15.9\std{0.2} & 43.1\std{0.3}  \\
RAW + Cityscapes-RAW (SRISP)    & 23.0\std{0.6} & 43.9\std{0.4} & 21.9\std{1.5} & 2.5\std{0.1} & 21.4\std{0.6} & 49.8\std{1.4}\\
RAW + BDD100K-RAW (SRISP)       & 24.2\std{0.3} & 45.7\std{0.4} & 22.6\std{0.7} & 2.9\std{0.3} & 21.7\std{0.3} & 51.8\std{1.2}\\
RAW + Cityscapes-RAW (ours)     & \underline{24.7}\std{0.3} & \underline{46.3}\std{0.6} & \underline{23.9}\std{0.2} & \underline{3.7}\std{0.3} & \underline{22.8}\std{0.6} & \underline{52.1}\std{0.9} \\
RAW + BDD100K-RAW (ours)        & \textbf{26.5}\std{0.3} & \textbf{49.3}\std{0.5} & \textbf{25.3}\std{0.5} & \textbf{4.4}\std{0.2} & \textbf{23.9}\std{0.4} & \textbf{54.6}\std{0.6} \\

\midrule
    & \multicolumn{6}{c}{\textbf{NOD Sony}} \\
    Training Dataset & $\operatorname{AP}$ & $\operatorname{AP_{50}}$ & $\operatorname{AP_{75}}$ & $\operatorname{AP_{\text{S}}}$ & $\operatorname{AP_{\text{M}}}$ & $\operatorname{AP_{\text{L}}}$ \\
\midrule
RGB         & 19.2\std{0.5} & 38.2\std{0.8} & 17.6\std{0.5} & 1.1\std{0.2} & 18.4\std{0.9} & 38.2\std{0.6} \\
RAW         & 18.0\std{0.1} & 35.8\std{0.3} & 16.3\std{0.4} & 1.1\std{0.3} & 16.6\std{0.1} & 36.8\std{0.3} \\
RAW + Cityscapes-RAW (SRISP)    & 23.1\std{0.4} & 46.9\std{0.4} & 20.0\std{0.1} & 2.0\std{0.2} & 21.3\std{1.0} & 44.4\std{0.1} \\
RAW + BDD100K-RAW (SRISP)       & 26.0\std{0.2} & 50.5\std{0.6} & 24.2\std{0.5} & 3.1\std{0.5} & 24.4\std{0.3} & 46.8\std{0.6} \\
RAW + Cityscapes-RAW (ours)     & \underline{26.2}\std{0.3} & \underline{50.6}\std{0.6} & \underline{24.9}\std{0.4} & \underline{3.4}\std{0.5} & \underline{25.4}\std{0.6} & \underline{46.9}\std{0.7} \\
RAW + BDD100K-RAW (ours)        & \textbf{28.6}\std{0.1} & \textbf{55.2}\std{0.4} & \textbf{26.4}\std{0.3} & \textbf{4.2}\std{0.1} & \textbf{27.7}\std{0.4} & \textbf{49.6}\std{0.3} \\
  \bottomrule
  \end{tabular}
\end{table*}

\begin{table*}[]
  \caption{Object detection results using YOLOv8 evaluating the performance on 100 NOD training samples (RGB and RAW) and the integration of Cityscapes-RAW and BDD100K-RAW generated by SRISP and RAW-Diffusion.\vspace{-0.1cm}}
  \label{tab:exp_results_object_detection_yolo_full}
  \small
  \centering
  \begin{tabular}{l|
        cccccc}
    \toprule
    & \multicolumn{6}{c}{\textbf{NOD Nikon}} \\
    Training Dataset & $\operatorname{AP}$ & $\operatorname{AP_{50}}$ & $\operatorname{AP_{75}}$ & $\operatorname{AP_{\text{S}}}$ & $\operatorname{AP_{\text{M}}}$ & $\operatorname{AP_{\text{L}}}$ \\
    \midrule
RGB & 22.7\std{1.7} & 32.9\std{10} & 22.5\std{1.8} & 1.7\std{0.3}  & 19.6\std{2.1} & 55.9\std{4.1}\\
RAW &  25.6\std{0.2}&	45.1\std{0.1}&	25.6\std{0.3}&	2.3\std{0.1}&	21.6\std{0.3}&	62.4\std{0.1} \\
RAW + Cityscapes-RAW (SRISP) &  29.1\std{0.1}&	48.6\std{0.2}&	29.4\std{0.1}&	3.1\std{0.1}&	25.3\std{0.1}&	66.0\std{0.4} \\
RAW + BDD100K-RAW (SRISP) &  \underline{32.0}\std{0.3}&	\underline{53.1}\std{0.5}&	\underline{32.1}\std{0.4}&	\underline{4.2}\std{0.2}&	\underline{28.6}\std{0.2}&	\underline{68.7}\std{0.5} \\
RAW + Cityscapes-RAW (ours) & 29.9\std{0.3}	&50.2\std{0.5}&	30.7\std{0.7}&	3.3\std{0.1}&	26.6\std{0.3}&	66.4\std{0.3} \\
RAW + BDD100K-RAW (ours) &  \textbf{32.6}\std{0.1}&	\textbf{54.3}\std{0.2}	& \textbf{32.6}\std{0.7}&	\textbf{4.9}\std{0.1}&	\textbf{29.0}\std{0.3}&	\textbf{69.2}\std{0.2} \\
\midrule
    & \multicolumn{6}{c}{\textbf{NOD Sony}} \\
    Training Dataset & $\operatorname{AP}$ & $\operatorname{AP_{50}}$ & $\operatorname{AP_{75}}$ & $\operatorname{AP_{\text{S}}}$ & $\operatorname{AP_{\text{M}}}$ & $\operatorname{AP_{\text{L}}}$ \\
\midrule
RGB & 18.4\std{0.3} & 33.8\std{0.2} & 18.4\std{0.4}& 1.9\std{0.0}  & 15.5\std{2.0} & 47.3\std{8.1} \\
RAW & 27.6\std{0.4}&	49.3\std{0.2}&	27.0\std{0.9}&	1.6\std{0.2}&	23.0\std{0.1}&	57.9\std{0.5} \\
RAW + Cityscapes-RAW (SRISP)&  29.5\std{0.4}&	51.7\std{0.6}&	29.1\std{0.7}&	2.9\std{0.3}&	25.2\std{0.2}&	58.0\std{1.6} \\
RAW + BDD100K-RAW (SRISP) &  \underline{32.0}\std{0.7}&	\underline{55.3}\std{0.8}&	\underline{31.6}\std{0.8}&	\underline{3.2}\std{0.0}&	\underline{29.5}\std{0.5}&	60.0\std{1.2} \\
RAW + Cityscapes-RAW (ours) & 31.0\std{0.3}&	54.8\std{0.5}&	31.0\std{0.5}&	\underline{3.2}\std{0.1}&	27.0\std{0.3}&	\underline{60.6}\std{0.4} \\
RAW + BDD100K-RAW (ours) & \textbf{33.6}\std{0.1}&	\textbf{58.3}\std{0.2}&	\textbf{33.8}\std{0.0}&	\textbf{4.2}\std{0.0}&	\textbf{31.0}\std{0.1}&	\textbf{61.9}\std{0.1} \\
  \bottomrule
  \end{tabular}
\end{table*}

\begin{table*}[]
  \caption{Zero-shot object detection results using Faster R-CNN. The models are trained exclusively on the generated datasets and evaluated on the NOD test set.\vspace{-0.1cm}}
  \label{tab:exp_results_object_detection_fasterrcnn_zeroshot_full}
  \small
  \centering
  \begin{tabular}{l|
        cccccc}
    \toprule
    & \multicolumn{6}{c}{\textbf{NOD Nikon}} \\
    Training Dataset & $\operatorname{AP}$ & $\operatorname{AP_{50}}$ & $\operatorname{AP_{75}}$ & $\operatorname{AP_{\text{S}}}$ & $\operatorname{AP_{\text{M}}}$ & $\operatorname{AP_{\text{L}}}$ \\
    \midrule
Cityscapes-RAW (SRISP) 
        & 7.7\std{1.0} & 16.8\std{2.7} & 6.4\std{0.6} & 0.4\std{0.2} & 9.5\std{0.7} & 23.8\std{1.4} \\
BDD100K-RAW (SRISP)  
        & \underline{18.4}\std{0.3} & \underline{35.9}\std{0.3} & \underline{15.8}\std{0.5} & \underline{2.5}\std{0.2} & \underline{16.7}\std{0.6} & \underline{38.1}\std{0.2} \\
Cityscapes-RAW (ours) 
        & 12.0\std{0.7} & 23.4\std{1.6} & 11.4\std{0.5} & 1.5\std{0.4} & 11.2\std{0.7} & 26.8\std{1.7}  \\
BDD100K-RAW (ours)          
        & \textbf{22.0}\std{0.1} & \textbf{43.1}\std{0.3} & \textbf{18.8}\std{0.5} & \textbf{3.5}\std{0.2} & \textbf{20.8}\std{0.2} & \textbf{43.4}\std{0.7} \\
\midrule
    & \multicolumn{6}{c}{\textbf{NOD Sony}} \\
    Training Dataset & $\operatorname{AP}$ & $\operatorname{AP_{50}}$ & $\operatorname{AP_{75}}$ & $\operatorname{AP_{\text{S}}}$ & $\operatorname{AP_{\text{M}}}$ & $\operatorname{AP_{\text{L}}}$ \\
\midrule
Cityscapes-RAW (SRISP) 
        & 3.8\std{1.1} & 9.3\std{2.8} & 2.6\std{0.5} & 0.2\std{0.1} & 4.7\std{0.6} & 14.1\std{1.7}\\
BDD100K-RAW (SRISP)  
        & \underline{16.9}\std{0.6} & \underline{34.2}\std{0.6} & \underline{14.5}\std{0.8} & .8\std{0.3} & \underline{15.6}\std{0.6} & \underline{30.2}\std{0.9} \\
Cityscapes-RAW (ours) 
        & 13.7\std{1.3} & 29.4\std{2.3} & 11.7\std{0.8} & \underline{1.9}\std{0.1} & 13.5\std{1.5} & 26.5\std{2.5} \\
BDD100K-RAW (ours)          
        & \textbf{21.6}\std{0.1} & \textbf{43.7}\std{0.4} & \textbf{18.7}\std{0.2} & \textbf{3.3}\std{0.2} & \textbf{20.9}\std{0.2} & \textbf{37.2}\std{0.4}\\
  \bottomrule
  \end{tabular}
\vspace{-0.0cm}  
\end{table*}

\begin{table*}[]
  \caption{Zero-shot object detection results using YOLOv8. The models are trained exclusively on the generated datasets and evaluated on the NOD test set.\vspace{-0.1cm}}
  \label{tab:exp_results_object_detection_yolo_zeroshot_full}
  \small
  \centering
  \begin{tabular}{l|
        cccccc}
    \toprule
    & \multicolumn{6}{c}{\textbf{NOD Nikon}} \\
    Training Dataset & $\operatorname{AP}$ & $\operatorname{AP_{50}}$ & $\operatorname{AP_{75}}$ & $\operatorname{AP_{\text{S}}}$ & $\operatorname{AP_{\text{M}}}$ & $\operatorname{AP_{\text{L}}}$ \\
    \midrule
Cityscapes-RAW (SRISP) 
        & 19.1\std{0.8} & 	33.5\std{1.2} & 	19.0\std{0.8} & 	1.7\std{0.2} & 	17.2\std{0.4} & 	48.9\std{2.0} \\
BDD100K-RAW (SRISP)  
        & 25.2\std{1.2} & 	43.9\std{2.6}	 & 24.3\std{1.1} & 	\underline{3.1}\std{0.2} & 	\underline{23.9}\std{1.3} & 	53.8\std{2.8} \\
Cityscapes-RAW (ours) 
        & \underline{25.8}\std{0.7} & 	\underline{44.1}\std{0.8} & 	\underline{25.7}\std{1.0} & 	2.7\std{0.2} & 	23.1\std{0.4} & 	\underline{58.8}\std{1.9}  \\
BDD100K-RAW (ours)          
        & \textbf{28.8}\std{0.6}	 & \textbf{49.8}\std{0.9} & 	\textbf{27.6}\std{0.9} & 	\textbf{4.3}\std{0.4} & 	\textbf{26.8}\std{0.1} & 	\textbf{59.9}\std{2.1} \\
\midrule
    & \multicolumn{6}{c}{\textbf{NOD Sony}} \\
    Training Dataset & $\operatorname{AP}$ & $\operatorname{AP_{50}}$ & $\operatorname{AP_{75}}$ & $\operatorname{AP_{\text{S}}}$ & $\operatorname{AP_{\text{M}}}$ & $\operatorname{AP_{\text{L}}}$ \\
\midrule
Cityscapes-RAW (SRISP) 
        & 17.4\std{1.0} & 	34.0\std{1.3}	 & 16.2\std{1.6} & 	1.8\std{0.1} & 	15.8\std{0.6} & 	39.7\std{2.3} \\
BDD100K-RAW (SRISP)  
        & 24.0\std{1.4} & 	43.6\std{2.7}	 & 23.1\std{1.3} & 	2.2\std{0.2} & 	22.8\std{1.6} & 	45.5\std{2.2} \\
Cityscapes-RAW (ours) 
        & \underline{26.1}\std{0.5} & 	\underline{48.8}\std{0.1} & 	\underline{25.0}\std{1.2} & 	\underline{2.6}\std{0.1} & 	\underline{23.4}\std{0.3} & 	\underline{53.6}\std{1.7} \\
BDD100K-RAW (ours)          
        & \textbf{29.2}\std{0.6} & \textbf{53.6}\std{1.0}	 & \textbf{28.1}\std{1.1} & 	\textbf{4.0}\std{0.3} & 	\textbf{27.3}\std{0.4} & 	\textbf{54.4}\std{1.9}\\
  \bottomrule
  \end{tabular}
\vspace{-0.0cm}  
\end{table*}

\begin{table*}[]
  \centering
  \caption{Analysis of integrating the original RGB dataset and our generated RAW dataset. The Average Precision (AP) is shown for each experiment.} 
  \label{tab:exp_results_od_rgb}
  \small
  \begin{tabular}{l|
        cc}
    \toprule
    Training Dataset & \multicolumn{1}{c}{Faster R-CNN} & \multicolumn{1}{c}{YOLOv8}\\
    \midrule
RAW                         & 18.2\std{0.2}     & 25.8\std{0.5} \\
RAW + Cityscapes-RGB        & 23.0\std{0.2}     & 26.1\std{0.3} \\ 
RAW + BDD100K-RGB           & 24.5\std{0.3}     & 27.3\std{0.3} \\
RAW + Cityscapes-RAW (ours) & 24.7\std{0.3} & 29.9\std{0.3} \\
RAW + BDD100K-RAW (ours)    & \textbf{26.5}\std{0.3} & \textbf{32.6}\std{0.1} \\
  \bottomrule
  \end{tabular}
\end{table*}

\end{document}